\documentclass[nohyperref]{article}

\usepackage[margin=1in]{geometry}
\usepackage[dvipsnames,svgnames, table,xcdraw]{xcolor}%

\usepackage{xspace}
\usepackage{graphicx}
\usepackage{color}
\usepackage{adjustbox}
\usepackage{booktabs}
\usepackage[numbers]{natbib}
\usepackage[colorlinks=true,linkcolor=blue,citecolor=blue,urlcolor=blue]{hyperref}
\hypersetup{
  colorlinks=true,
  citecolor=Blue,
  linkcolor=BrickRed,
} 

\usepackage{algorithm}
\usepackage{algpseudocode}
\usepackage{url}
\usepackage{comment}
\usepackage{amsmath}
\usepackage{bbm}
\usepackage{booktabs}
\usepackage{array}
\usepackage{pgfplots}
\usepackage{geometry}
\usepackage{adjustbox}
\usepackage{graphicx}
\usepackage{colortbl}
\usepackage{hyperref}
\usepackage[toc,page]{appendix}
\usepackage[most]{tcolorbox}
\usepackage{tikz}
\usetikzlibrary{arrows.meta}  
\usetikzlibrary{backgrounds}  
\usetikzlibrary{positioning}   
\usetikzlibrary{shapes}        
\usetikzlibrary{arrows.meta}   
\usepackage{booktabs}
\usepackage{array}
\usepackage{multirow}
\usepackage{tabularx}
\usepackage{tikz}
\usepackage{hyperref}

\definecolor{headercolor}{RGB}{46, 116, 181}
\definecolor{row1}{RGB}{235, 241, 252}
\definecolor{row2}{RGB}{255, 255, 255}
\definecolor{categorycolor}{RGB}{100, 149, 237}
\definecolor{linkcolor}{RGB}{0, 102, 204}

\hypersetup{
    colorlinks=true,
    linkcolor=linkcolor,
    urlcolor=linkcolor
}

\usepackage{longtable}
\newcolumntype{X}{>{\centering\arraybackslash}m{3cm}}
\newcolumntype{Y}{>{\centering\arraybackslash}m{4cm}}
\newcolumntype{Z}{>{\centering\arraybackslash}m{9cm}}

\usepackage[normalem]{ulem}
\usepackage{xcolor}

\newcommand{\scrcmd}{\textsuperscript}
\newcommand{\scr}[1]{\scrcmd{#1}}

\title{Can LLM feedback enhance review quality? A randomized study of 20K reviews at ICLR 2025}
\author{Nitya Thakkar$^{1}$, Mert Yuksekgonul$^{1}$, Jake Silberg$^{1}$, Animesh Garg$^{2}$, \\ Nanyun Peng$^{3}$, Fei Sha$^{4}$, Rose Yu$^{5}$, Carl Vondrick$^{6}$, James Zou$^{1}$\\
\normalsize{$^{1}$Stanford University}\\
\normalsize{$^{2}$Georgia Institute of Technology}\\
\normalsize{$^{3}$University of California, Los Angeles}\\
\normalsize{$^{4}$Google Research}\\
\normalsize{$^{5}$University of California, San Diego}\\
\normalsize{$^{6}$Columbia University}\\
}
\date{}

\begin{document}
\maketitle

\begin{abstract}
Peer review at AI conferences is stressed by rapidly rising submission volumes, leading to deteriorating review quality and increased author dissatisfaction. To address these issues, we developed Review Feedback Agent, a system leveraging multiple large language models~(LLMs) to improve review clarity and actionability by providing automated feedback on vague comments, content misunderstandings, and unprofessional remarks to reviewers. Implemented at ICLR 2025 as a large randomized control study, our system provided optional feedback to more than 20,000 randomly selected reviews. To ensure high-quality feedback for reviewers at this scale, we also developed a suite of automated reliability tests powered by LLMs that acted as guardrails to ensure feedback quality, with feedback only being sent to reviewers if it passed all the tests. The results show that 27\% of reviewers who received feedback updated their reviews, and over 12,000 feedback suggestions from the agent were incorporated by those reviewers. This suggests that many reviewers found the AI-generated feedback sufficiently helpful to merit updating their reviews. Incorporating AI feedback led to significantly longer reviews (an average increase of 80 words among those who updated after receiving feedback) and more informative reviews, as evaluated by blinded researchers. Moreover, reviewers who were selected to receive AI feedback were also more engaged during paper rebuttals, as seen in longer author-reviewer discussions. This work demonstrates that carefully designed LLM-generated review feedback can enhance peer review quality by making reviews more specific and actionable while increasing engagement between reviewers and authors. The Review Feedback Agent is publicly available at \href{https://github.com/zou-group/review_feedback_agent}{https://github.com/zou-group/review\_feedback\_agent}.
\end{abstract}

\section{Introduction}
\label{sec:intro}
Scientific peer review is a critical step before publication, where domain experts evaluate the research to ensure thoroughness and scientific integrity, prevent false claims, and provide a strong foundation for future work \cite{87affd2d-c6eb-361c-b98f-e76719a9bec4, kelly2014peer}. High-quality reviews are essential for authors to improve their work, address key limitations, and advance scientific progress. However, in a survey of 11,800 researchers worldwide, while 98\% view peer review as essential to maintaining the quality and integrity of academic communication, only 55.4\% expressed satisfaction with the quality of reviews they receive \cite{Publons2018}. This dissatisfaction has grown as obtaining constructive and high-quality peer reviews has become more challenging due to the increase in the number of paper submissions, especially in fast-moving areas like Artificial Intelligence (AI) \cite{azad2024publicationtrendsartificialintelligence, mccook2006}. For example, the International Conference on Learning Representations (ICLR) experienced year-over-year submission increases of 47\% in 2024 and 61\% in 2025 \cite{iclr2024pressrelease}. To maintain a rigorous and meaningful peer review process amid this growth, it is crucial to address the growing burden on reviewers and the subsequent deterioration in review quality. 

Authors at AI conferences increasingly report receiving short, vague reviews with criticisms like `not novel' or `not state-of-the-art (SOTA)' \cite{rogers-augenstein-2020-improve}. At the 2023 Association for Computational Linguistics meeting, authors flagged 12.9\% of reviews for poor quality, primarily due to these vague, surface-level criticisms \cite{rogers2023program}. The peer review system is further strained by reviewers being assigned papers outside their expertise \cite{Arns2014} and the same papers being reviewed multiple times due to high rejection rates \cite{87affd2d-c6eb-361c-b98f-e76719a9bec4}. Additionally, the 2014 NeurIPS Experiment highlighted inconsistencies in the peer review process by showing that approximately 25\% of paper acceptance decisions differed between two independent review committees \cite{cortes2021inconsistencyconferencepeerreview}. These issues not only frustrate authors but potentially allow weaker research to be accepted while strong work is rejected, ultimately preventing papers from reaching their full potential due to the decline of meaningful dialogue between reviewers and authors.

Large language models (LLMs) \cite{claude} have the potential to enhance the quality and usefulness of peer reviews for authors ~\citep{liang2024can}. Recent studies demonstrated that LLMs can serve as effective critics, generating detailed and constructive feedback \cite{yuksekgonul2024textgradautomaticdifferentiationtext, madaan2023selfrefineiterativerefinementselffeedback}. Furthermore, LLMs have already shown high utilization in the peer review process. Reviewers are increasingly turning to LLMs to assist in drafting their reviews, with an estimated 10.6\% of reviewers at ICLR 2024 using LLMs for this purpose \cite{Hosseini2023, liang2024monitoringaimodifiedcontentscale}.  

To explore how LLMs can improve review quality at scale, we introduce \emph{Review Feedback Agent}, a multi-LLM system designed to enhance the clarity and actionability of reviews by providing feedback to reviewers. Piloted at ICLR 2025 as a large randomized control study, our agent provided feedback to over 20,000 randomly selected reviews (representing half of all ICLR 2025 reviews) over four weeks from October 15 to November 12, 2024. The generated feedback primarily focused on minimizing instances of vague and unjustified comments while also addressing content misinterpretations and unprofessional remarks. Using Claude Sonnet 3.5 as the backbone \cite{claude}, we created a system of five LLMs that collaborated to generate high-quality feedback. To enhance the system's reliability against potential errors or failures in instruction-following ~\citep{zhang2023siren, zhou2023instruction}, we developed a set of reliability tests to evaluate specific qualities of the generated feedback; the feedback was only posted if it passed all of these tests.

\textbf{Summary of main findings.} Of the randomly selected ICLR reviews that received AI feedback, 26.6\% of reviewers updated their reviews, altogether incorporating 12,222 suggestions from the feedback agent into the reviews. Blinded ML researchers labeled these revised reviews as more informative and clearer than their initial versions. Reviewers who updated after receiving feedback increased the length of reviews by an average of 80 words. Furthermore, AI feedback led to more engaged discussions during the rebuttal period, as seen through longer author and reviewer responses. We also observed that reviewers who received feedback were more likely to change their scores after the rebuttal period, which was consistent with a more engaged rebuttal process. 

In this study, we present the first large-scale deployment for using LLMs to assist peer review. By making reviews more actionable and informative, we aim to enhance the peer review experience and promote a more constructive scientific process. 

\section{Methods}
\label{sec:methods}
In what follows, we first describe the review feedback experiment, including its goals and our technical setup with OpenReview. Next, we outline the architecture of our Review Feedback Agent and explain how the system was designed to meet our goals while ensuring a high level of reliability. In total, the agent automatically provided feedback to over $20,000$ reviews at ICLR 2025.

\begin{figure}[hbtp]
    \centering
    \includegraphics[width=\textwidth]{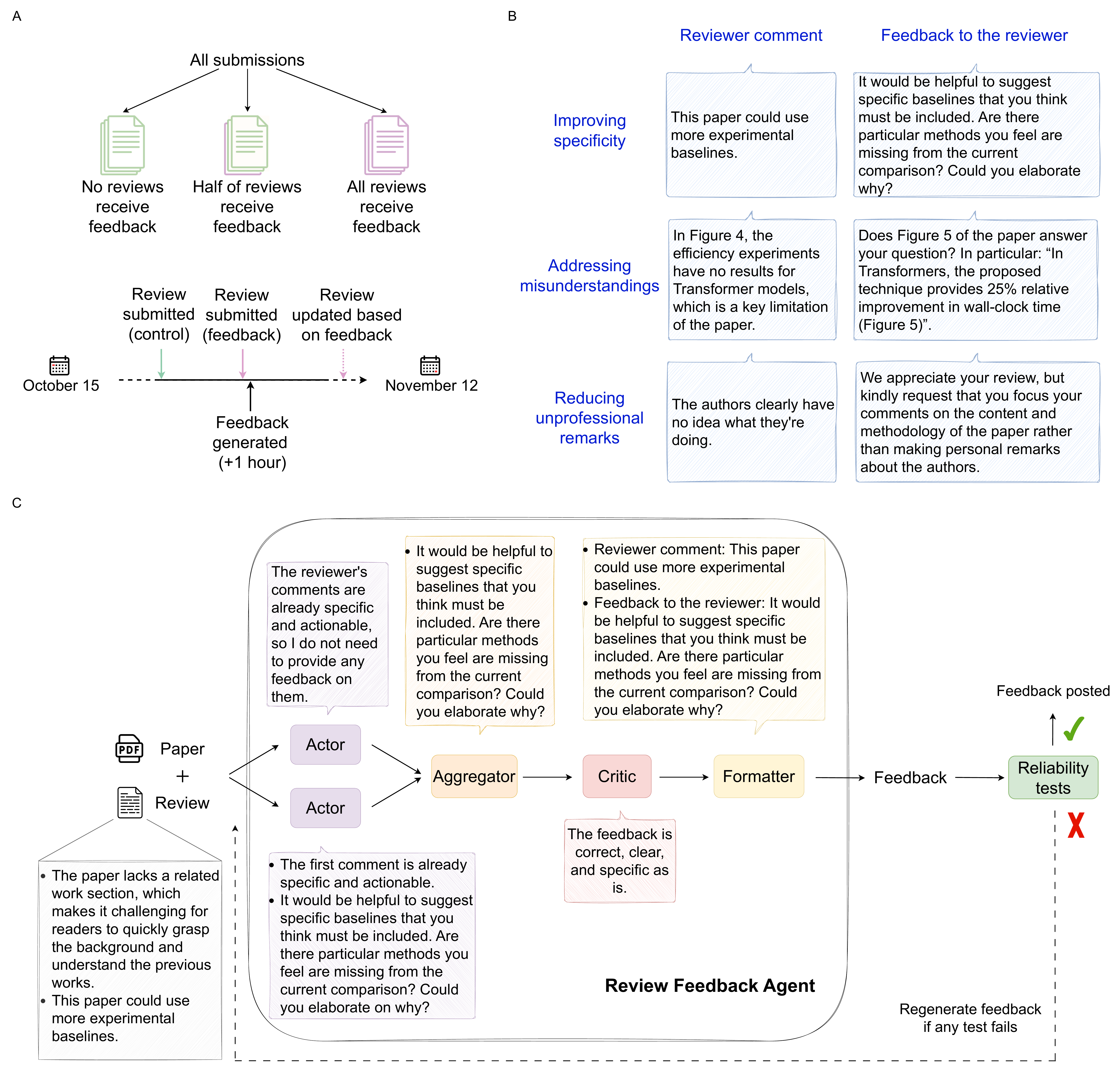}
    \caption{\textbf{(A) Randomized controlled study setup.} Before the start of the review period, we randomly assigned all submissions to one of three groups to determine how many of its reviews received feedback: none, half, or all. When a review selected to receive feedback was submitted, the agent generated and posted feedback after 1 hour. Reviewers could update their review, optionally, based on the feedback until the end of the review period, which ran from October 14 to November 12, 2024. \textbf{(B) Feedback categories.} Our system is designed to address three main types of review comments. Here, we provide examples of comments that would receive feedback from our agent, as well as examples of the generated feedback. \textbf{(C) Review Feedback Agent.} Our system consists of five LLMs ~(\emph{Actors, Aggregator, Critic, and Formatter}). Two parallel \emph{Actor}s generate the initial feedback, then pass it to the \emph{Aggregator}, the \emph{Critic}, and finally the \emph{Formatter}. Finally, the feedback is passed through the reliability tests; upon successfully passing, the feedback is posted on a review. We provide examples of comments and feedback given to those comments by our system.}
    \label{fig:figure1}
\end{figure}

\subsection{ICLR 2025 review feedback experiment}

Our pilot study was conducted in collaboration with ICLR 2025 and OpenReview. As one of the world's fastest-growing AI conferences, ICLR receives thousands of paper submissions yearly; in 2025, ICLR received 11,603 submissions. Each submission is assigned an average of 4 reviewers, and all reviews are standardized to include the same sections: summary, strengths, weaknesses, and questions. Furthermore, reviewers provide scores on a scale of 1 (low) to 10 (high), rating the paper according to the following categories: soundness, presentation, contribution, rating, and confidence.

\textbf{Goal:} Our goal was to enhance review quality and, in particular, reduce low-information content reviews. Toward this goal, we identified three categories of common issues in reviews that we hoped to improve by providing LLM-generated feedback. The common issues are: 1) vague or generic critiques in reviews (the feedback asks the reviewers to be more specific and actionable); 2) questions or confusions that could be addressed by overlooked parts of the paper (the feedback highlights relevant sections); and 3) unprofessional statements in the review (the feedback asks the reviewer to rephrase). For each comment in a review, the Review Feedback Agent determined if it fell into any of these problematic categories and, if so, provided feedback on that specific review comment.

\textbf{Experimental setup:} We set up this experiment as a Randomized Control Trial (RCT) to enable us to make causal inferences about how receiving feedback influences the peer review process. Before the beginning of the review period, we randomly split papers into one of three equal groups (see Figure \ref{fig:figure1}A):
\begin{enumerate}
    \item No reviews for this paper will receive feedback,
    \item Half of the reviews for this paper will be randomly selected to receive feedback,
    \item All reviews for this paper will receive feedback.
\end{enumerate}

\begin{figure}[hbtp]
    \centering
    \includegraphics[width=0.7\textwidth]{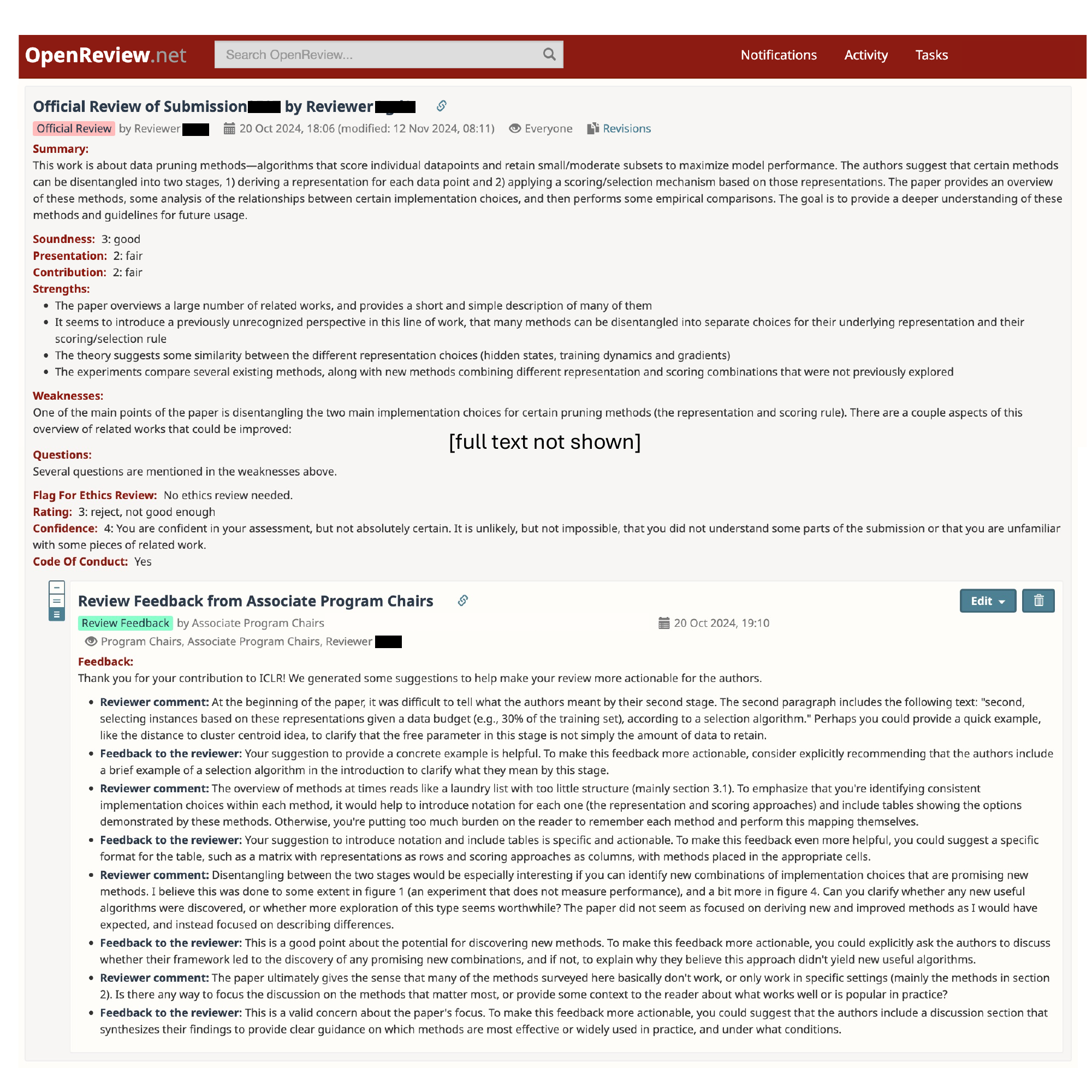}
    \caption{\textbf{OpenReview interface.} Here, we provide an example of feedback posted to a review on the OpenReview website (with consent from the reviewer). Feedback is only visible to the reviewer and the ICLR program chairs and was posted roughly one hour after the initial review was submitted.}
    \label{fig:figure2}
\end{figure}

For reviews randomly assigned to receive feedback, the Review Feedback Agent, wrapped in an API, was automatically triggered when a reviewer first submitted their review on OpenReview. We delayed the feedback generation by one hour after a review was initially submitted to allow reviewers time to make any small edits (e.g., typo corrections). See Figure \ref{fig:figure1}A for an example timeline. The agent posted feedback to reviews through the OpenReview interface by replying to reviews with the feedback wrapped in a comment. See Figure \ref{fig:figure2} for an example of what feedback looked like on the OpenReview website.

The agent only provided feedback on the initial review, and there was no subsequent interaction between the reviewer and the feedback system after that time point. The feedback is only visible to the reviewer and the ICLR program chairs; it was not shared with other reviewers, authors, or area chairs and was not a factor in the acceptance decisions. Reviewers were informed that the feedback was generated by a LLM and could choose to ignore the feedback or revise their review in response, as the system did not make any direct changes. Finally, we did not access or store any identifiable information about authors or reviewers. This study was reviewed by IRB and deemed low risk.

\textbf{Statistics:} Around 50\% of reviews were randomly selected to receive feedback. Of the 44,831 reviews submitted on 11,553 unique papers (we excluded desk-rejected submissions), we posted feedback to 18,946 reviews (42.3\%) over 4 weeks from October 15 to November 12, 2024 (see Figure \ref{fig:figure2}A). Less than 8\% of the selected reviews did not receive feedback for one of two reasons: 2,692 reviews were originally well-written and did not need feedback, while 829 reviews had feedback that failed the reliability tests. Each review took roughly one minute to run through our entire pipeline and cost around 50 cents. On average, each review that received feedback was given 3-4 feedback comments, with a minimum of 1 and a maximum of 17.

\subsection{Review Feedback Agent}

The Review Feedback Agent aimed to provide feedback that helped reviewers make their comments more specific, constructive, and actionable for the authors. 

\textbf{Feedback categories:} The Review Feedback Agent provided suggestions on three potential categories of issues in reviews. We curated these categories by examining reviewer guidelines from several AI conferences \cite{icml2023reviewertutorial, icml2022reviewertutorial, acl2017revieweradvice, Valdenegro2021reviews} and evaluating previously identified patterns of ``lazy reviewer thinking'' \cite{rogers-augenstein-2020-improve}. We also took inspiration from the ARR guidelines, where 16 common reviewer heuristics are outlined \cite{arrguidelines}. Importantly, the agent was not designed to suggest new ideas to add to the review; rather, it only focused on revising the existing ideas and preventing lower-quality reviews. The target feedback areas that we ultimately focused on were:

\begin{enumerate}
    \item \textbf{Improving specificity:} Encouraging reviewers to rephrase vague review comments, making them more specific, actionable, and justified for the authors.
    \item \textbf{Addressing misunderstandings:} Highlighting sections of the paper that may already address some of the reviewer’s questions or confusion.
    \item \textbf{Reducing unprofessional remarks:} Identifying and addressing unprofessional or inappropriate remarks in the review. A 2019 study of 1,106 researchers found that 58\% had received an unprofessional review, highlighting its prevalence \cite{silbiger2019unprofessional}.
\end{enumerate}

See Figure \ref{fig:figure1}B for examples of real reviewer comments (from ICLR 2024 reviews and public journal reviews) in each category that would receive feedback and examples of feedback that would be given. 

\textbf{Preprocessing:} The agent was provided with the paper PDF's text (extracted using pypdf's PdfReader \cite{pypdf}) and the review text as input. We extracted the summary, strengths, weaknesses, and questions sections from the review. We did not provide the agent with any of the scores the reviewer initially gave the paper.

\textbf{Architecture:} The agent generated a list of pairs, with each pair consisting of a review comment that fit into one of the problematic categories above and the corresponding feedback provided for that comment. The agent was composed of a pipeline of five LLMs (see Algorithm \ref{alg:agent}, Figure \ref{fig:figure1}C). We used the Claude Sonnet 3.5 (June 20, 2024) model \cite{claude} as the backbone; we picked the backbone model by generating feedback with the same prompt using GPT-4o, Gemini 1.5 Flash, and Claude Sonnet 3.5 and then conducting a blind preference evaluation. Additionally, through testing, we found that one LLM was insufficient to generate high-quality feedback and format it correctly, thus, we instantiated the multi-call pipeline. First, we defined two parallel actor LLMs to generate the initial set of feedback based on the previously defined target areas. The actors were provided with the initial review ($R$) and paper text ($P$) as inputs. We used two separate actors to optimize for feedback diversity. Then, we passed the two lists of feedback, $F_1$ and $F_2$, to an aggregator LLM, which merged the lists into one set of feedback, $F_{combined}$. Next, we passed this candidate list to a critic LLM responsible for ensuring the feedback was accurate and clear. Importantly, the critic also removed any feedback that was too superficial or nitpicky, defined through various in-context examples (see Appendix \ref{sec:agent_prompts} for the examples), as we did not want to overwhelm or annoy reviewers. Finally, a formatter LLM was provided with this final list, $F_{filtered}$, and formatted it into pairs:
\begin{itemize}
    \item **Reviewer comment:** {{a comment}}
    \item **Feedback to the reviewer:** {{feedback to the comment}}
\end{itemize}

\begin{algorithm}
\caption{Review Feedback Agent}
\label{alg:agent}
\begin{algorithmic}[1]
    \State \textbf{Input:} Paper text $P$, Review $R$, max attempts $T = 2$
    \For{$t = 1$ to $T$}
        \State $F_1 \gets \text{Actor}_1(P, R)$
        \State $F_2 \gets \text{Actor}_2(P, R)$
        
        \State $F_{combined} \gets \text{Aggregator}(F_1, F_2)$
        \State $F_{filtered} \gets \text{Critic}(F_{combined})$
        \State $F_{final} \gets \text{Formatter}(F_{filtered})$
        
        \If{PassReliabilityTests($F_{final}$)}
            \Return $F_{final}$
        \EndIf
    \EndFor
    \State \textbf{return} error
\end{algorithmic}
\end{algorithm}

See Appendix \ref{sec:agent_prompts} for the exact prompts used. To refine this system, we constructed a test set of 50 ICLR 2024 reviews we perceived to be of low quality in one or more of our target areas (i.e., they made vague comments, asked questions that were present in the paper already, and/or made unprofessional remarks). We iteratively ran our agent on this test set, examined the generated feedback, and refined the prompts to optimize the results. This procedure ultimately led to prompts that produced high-quality feedback for all 50 reviews in the test set.

\textbf{Reliability testing:} Inspired by \cite{ptest}, we also developed a suite of reliability tests designed to act as guardrails, ensuring the quality of our generated feedback. Reliability tests evaluate specific attributes of a model’s output. The four reliability tests we developed ensured the feedback provided constructive suggestions, addressed the reviewer, did not simply restate what the reviewer wrote, and was formatted correctly. We provide the exact reliability tests we used and examples of feedback that would fail the reliability tests in Appendix \ref{sec:reliability_tests}. We developed up to five test cases for each reliability test and refined the reliability test prompts until we passed all the test cases. To refine our Review Feedback Agent's pipeline and prompts, we passed our test set reviews through the validated reliability tests until we achieved a 100\% pass rate.

Feedback was only posted to a review if it passed all our reliability tests; if it failed, we re-ran the entire pipeline a second time ($T=2$) to generate new feedback. Upon a second fail, we returned an error and did not post the feedback. Over 96\% of generated feedback for ICLR 2025 reviews passed all reliability tests.

\section{Results}
\label{sec:results}
\subsection{Impact of feedback on review updates and reviewer engagement}

\begin{figure}[hbtp]
    \centering
    \includegraphics[width=0.9\textwidth]{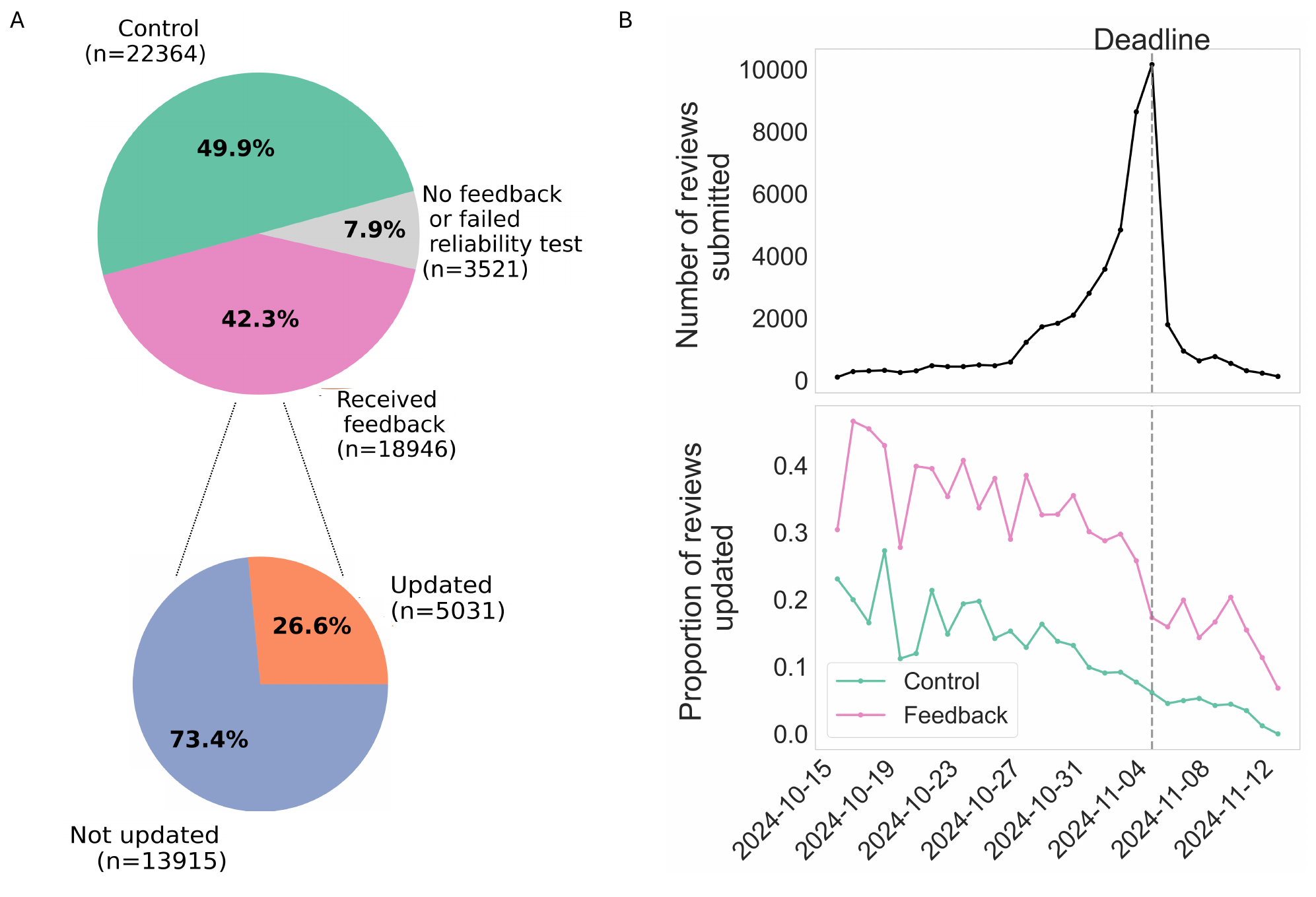}

    \begin{center}
    \begin{minipage}{0.9\textwidth}
    \raggedright
    \footnotesize C
    \end{minipage}
    \end{center}

    \begin{minipage}{0.9\textwidth}
    \centering
    \begin{tabular}{lcccc}
    \toprule
    \textbf{} & \textbf{Control} & \textbf{Feedback} & \textbf{Not updated} & \textbf{Updated}  \\
    \midrule
    Average change in length & $7.0$ & $21.0^{**}$ ($\Delta + 200\%$) & $2.1$ & $80.3^{***}$ \\
    \bottomrule
    \end{tabular}
    \end{minipage}
    
    \caption{\textbf{(A) Feedback statistics.} Among all ICLR 2025 reviews, 22,467 were randomly selected to receive feedback (feedback group), and 22,364 were randomly selected not to receive feedback (control group). Of those selected to receive feedback, 18,946 (42.3\%) successfully received feedback, with 26.6\% of those reviewers updating their reviews. \textbf{(B) Update rates.} (Top) Most reviews were submitted 2-3 days before the review deadline (November 4, 2024). (Bottom) Reviewers were more likely to update their review if they submitted it early relative to the deadline. Reviewers who received feedback were much more likely to update their reviews than those in the control group, with a difference of approximately 17 percentage points. \textbf{(C) Average change in review length} (measured as number of words). Review length is measured only for the following sections: summary, strengths, weaknesses, and questions. The difference in review length between the control and feedback groups is statistically significant ($^{**}$p $\leq$ 0.01), with being selected to receive feedback leading to an average increase of 14 words more (a 200\% increase) in review length compared to the control group. The difference is more pronounced between the not-updated and updated groups ($^{***}$p $\leq$ 0.001).}
    \label{fig:figure3}
\end{figure}

First, we aimed to objectively measure how many reviewers updated their reviews after receiving feedback compared to those who did not receive feedback. This enabled us to assess how the feedback may have been associated with changes in various components of their review, such as length and scores.

We conducted this ICLR experiment as a randomized controlled study by randomly splitting all reviews into one of two groups: not selected to receive feedback (\emph{control group}) or selected to receive feedback (\emph{feedback group}) - see Section \ref{sec:methods} for more details. Note that the group selected to receive feedback includes the 7.9\% of reviews that were selected but did not actually receive feedback, mostly because AI deemed feedback not necessary there. This intent-to-treat definition of the feedback group enables us to conduct causal analysis but could dilute the actual effect of the feedback.  

Of all reviews in the feedback group, we further defined reviews that successfully received feedback as either being \emph{not updated} or \emph{updated}. A review is \emph{not updated} if a reviewer did not edit their review after receiving feedback or if the edit distance between the initial and modified review was less than 5; this edit distance filtering accounted for minor updates such as fixing typos or modifying scores. Conversely, a review is \emph{updated} if a reviewer did edit their review after receiving feedback and the edit distance between the initial and modified review was greater than 5.

Of the 18,946 reviews that successfully received feedback, 5,031 (26.6\%) reviews were updated (Figure \ref{fig:figure3}A). Out of the 22,364 reviews in the control group, only 2,103 (9.4\%) were updated; here, we define updated for the control group as a reviewer updating at least one hour after posting (the time it takes for the feedback group to receive feedback) with an edit distance greater than 5. With an update rate difference of roughly 17 percentage points (Figure \ref{fig:figure3}B), we can see that reviews that received feedback were much more likely to be updated than those that did not. 

In Figure \ref{fig:figure3}B, we also see that reviewers who submitted early relative to the deadline (November 4, 2024) were more likely to update their review than those who submitted close to or after the deadline. This suggests that more organized reviewers, who may already be more engaged in the review process, were more likely to revise their reviews in response to feedback. While this will influence our analysis comparing the not updated and updated groups, we can be confident that the underlying distribution of the control and feedback groups is similar and not biased by factors such as reviewer organization because we conducted this as an RCT. Randomization helps mitigate such biases, making it possible to assess the causal impact of the feedback on the peer review process.

Finally, we analyzed the change in review length (number of words in the summary, strengths, weaknesses, and questions sections) among the groups (Figure \ref{fig:figure3}C). We compared the initial review length and the modified review length; we refer to modified reviews as the review at the end of the four-week review period before the rebuttal period began (only these modified reviews are made public to authors). We saw that review length, on average, increased across all groups. First, we observed that being selected to receive feedback caused the average review length to increase by about 14 words more than reviews that were not selected to receive feedback. Note that this effect size is deflated due to the substantial number of reviewers who received feedback but did not update their review, as well as the 7.9\% of reviews that were selected to receive feedback but did not actually get it. We also see that updating the review after receiving feedback is associated with a statistically significant increase in review length (80 words) compared to not updating the review (2 words). We can infer that reviewers who updated their reviews were editing them more consistently to incorporate more detail and nuance, explaining this large increase in length. In addition to feedback causing an increase in review length, we also found that a significantly higher percentage of reviewers who received feedback edited at least one of their scores (soundness, presentation, contribution, rating, and confidence) during the review period, with 8.1\% of them making edits compared to 7.5\% among the control group (p $\leq$ 0.05). In Appendix \ref{sec:score_changes},  we observe no significant difference in the average score changes between the feedback and control groups.

\subsection{Measuring how much feedback reviewers incorporate}

Of the reviewers that updated their review, we wanted to measure what proportion of them incorporated one or more pieces of feedback they were provided. This analysis helped us estimate how many reviewers found the feedback useful.

\begin{figure}[hbtp]
    \centering
    \includegraphics[width=0.9\textwidth]{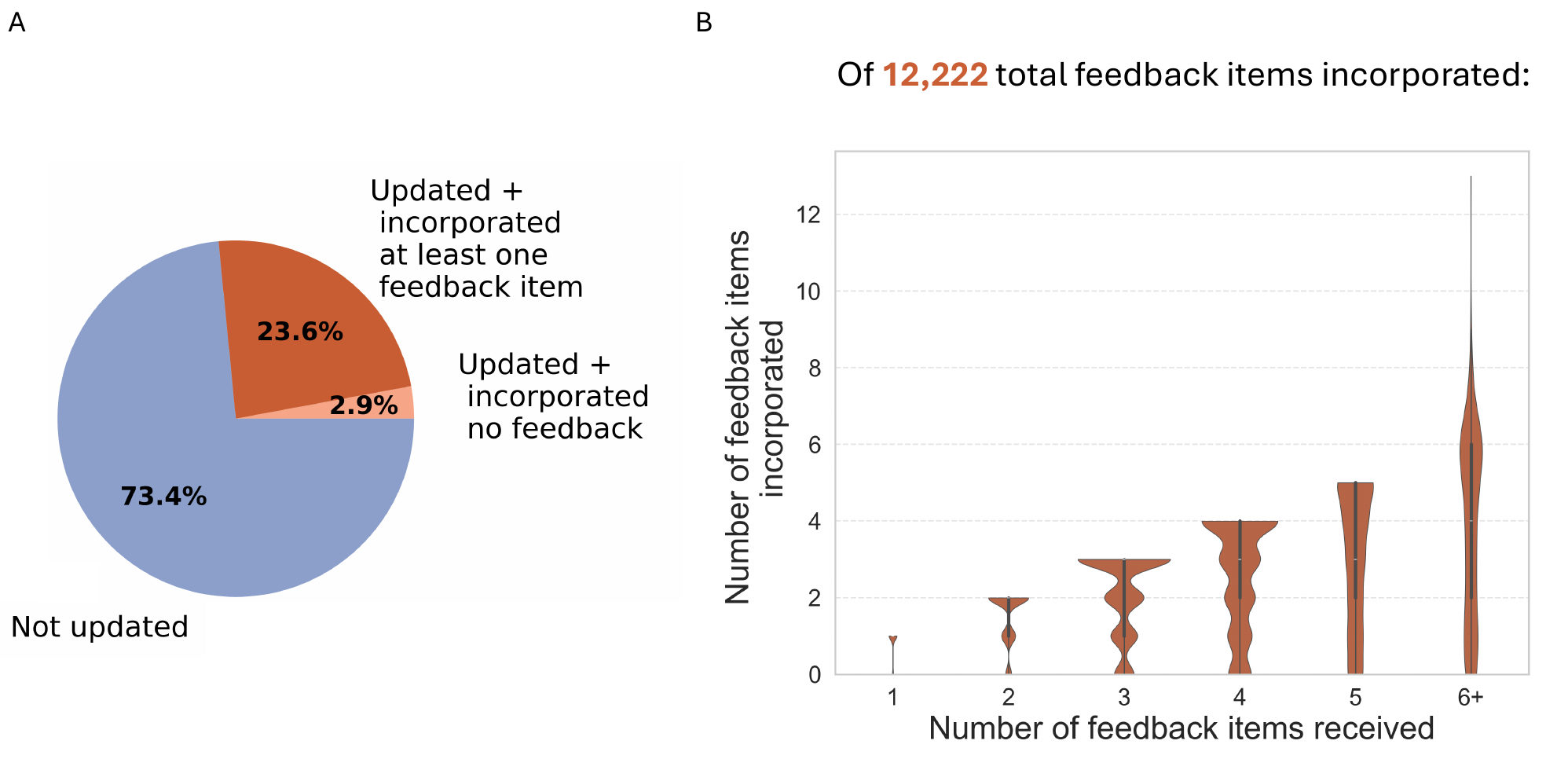}
    \caption{\textbf{(A) Overall incorporation statistics.} Through our LLM-based incorporation analysis, we estimate that 23.6\% of reviewers who were given feedback incorporated at least one feedback item they were given. This means that 89\% of reviewers who updated their review after receiving feedback incorporated at least one item. \textbf{(B) Feedback incorporation trends.} Here, we illustrate the relationship between the number of feedback items reviewers who updated their review received and how many of those items they incorporated. In total, reviewers incorporated 12,222 feedback items. Notably, reviewers were more likely to incorporate feedback when given fewer items.}
    \label{fig:figure4}
\end{figure}

We counted a piece of feedback as incorporated if the reviewer clearly integrated some part of the feedback into their modified review. To systematically carry out this analysis, we developed an LLM-based pipeline to run on all updated reviews (see Supplementary Figure \ref{fig:supp_conf_matrix}A). We used the Claude Sonnet 3.5 model to evaluate whether each feedback item received by a reviewer was incorporated into their modified review. See Appendix \ref{sec:incorp_validation} for our approach to validating this pipeline. Of the 5,031 reviews that reviewers updated, encompassing 18,322 total feedback items, 89\% of reviewers incorporated at least one piece of feedback. This represents 23.6\% of all reviewers who received feedback (Figure \ref{fig:figure4}A). In total, we estimate that 12,222 feedback items were incorporated into revised reviews. We also examined the number of feedback items reviewers who updated their reviews received compared to how many feedback items they incorporated (Figure \ref{fig:figure4}B). We see that when reviewers receive fewer feedback items, they are more likely to incorporate more (or even all) of the items. Overall, the average reviewer who updated their review incorporated 69.3\% of the feedback they received; in other words, given 3 pieces of feedback, the average reviewer who updated their review incorporated 2 of them.

Below are a few examples of real incorporations reviewers made based on their feedback. We provide the initial review comment they posted, the feedback they received, and then their modified comment.

\begin{tcolorbox}[colback=blue!5!white,colframe=blue!75!black,title=Incorporation example: Feedback that encourages specificity]

\textbf{Original comment:} Theorems 1, 2, 3 require the notion of interpolation, which is considered a strong assumption to have. 

\vspace{1em}

\textbf{Feedback:} To make this critique more actionable, it would be helpful to elaborate on why the interpolation assumption is problematic in this context. Are there particular types of federated learning problems where this assumption is unlikely to hold? Providing concrete examples or suggestions for how the authors could relax or justify this assumption would give them clearer guidance on addressing this limitation. 

\vspace{1em}

\textbf{Revised comment:} Theorems 1, 2, 3 require the notion of interpolation. Even though an explanation of regimes that satisfy this condition is provided, considering that there are previous works [1], [2] that extend beyond that setting, this assumption seems to be an avenue for future work in this field. More specifically, the initial FedProx algorithm [1] is analyzed in the general non-interpolated setting. In addition, the follow-up work regarding the FedExProx algorithm [2] considers in the main paper the interpolated regime. However, the authors provide additionally an illustration of the algorithm's behaviour in the non-interpolated setting (see Appendix F.3 in [2]). In that sense, it would be useful to provide some additional details on the behaviour of the algorithm in the non-interpolated setting or to comment on the main challenges in extending the current proof technique beyond the interpolation framework, offering in that way a more complete picture and direction for future research.
  
\end{tcolorbox}

\begin{tcolorbox}[colback=blue!5!white,colframe=blue!75!black,title=Incorporation example: Feedback that identifies content oversight]

\textbf{Original reviewer comment:} How is the performance on Llama?

\vspace{1em} 

\textbf{Feedback to the reviewer:} The paper appears to provide results on Llama 2-7B in Table 3. Does this address your question? If not, could you clarify what specific aspects of performance on Llama you're interested in that aren't covered by the current results?

\vspace{1em}

\textbf{Modified reviewer comment:} How is the performance on Llama3 with the alpaca dataset?
  
\end{tcolorbox}

\begin{tcolorbox}[colback=blue!5!white,colframe=blue!75!black,title=Incorporation example: Feedback that makes review more actionable]

\textbf{Original reviewer comment:} Limited application focus on other visual tasks (e.g., object detection, segmentation) beyond classification.

\vspace{1em}

\textbf{Feedback to the reviewer:} To make this comment more actionable, consider asking the authors to discuss potential extensions or challenges in applying their approach to other visual tasks. \emph{For example: `How might the visual memory approach be adapted for tasks like object detection or segmentation? What unique challenges would arise in those domains?' }

\vspace{1em}

\textbf{Modified reviewer comment:} Limited application focus on other visual tasks (e.g., object detection, segmentation) beyond classification. \emph{How might the visual memory approach be adapted for tasks like object detection or segmentation? What unique challenges would arise in those domains?}
  
\end{tcolorbox}

Finally, we wanted to assess whether these incorporated reviews were clearer, more specific, and more actionable for authors. To conduct this analysis, we asked two human AI researchers to conduct a blind preference evaluation between the initial and modified pre-rebuttal reviews. Specifically, we focused on reviews in the updated group that received 3-4 feedback items (the average number given) where the proportion of incorporated feedback exceeded 0.60. This threshold was chosen because the average incorporation rate was 67\%, and we aimed to assess whether an average updated review with incorporated feedback was perceived as an improvement. Human annotators preferred modified reviews 89\% of the time (out of 100 examples), indicating that reviewers who incorporated feedback consistently produced higher-quality reviews.

\subsection{Influence of feedback on rebuttals and decisions}

We next analyzed the impact of being selected to receive feedback on the rebuttal process and decision outcomes. The rebuttal period took place over three weeks between November 12 and December 4, 2024, and was a time when authors could respond to their reviewer's comments as they revised their papers. We examined how the feedback causally impacted different engagement measures during the rebuttal period.

\begin{table}[h!]
\centering
\begin{tabular}{lcccc}
\toprule
\textbf{} & \textbf{Control} & \textbf{Feedback} & \textbf{Not updated} & \textbf{Updated}  \\
\midrule
Average length of author rebuttal & $807$ & $855^{***}$ ($\Delta$ + 6\%) & $840$ & $896^{***}$ \\
Average length of reviewer replies & $110$ & $116^{***}$ ($\Delta$ + 5.5\%) & $115$ & $129^{***}$\\
\bottomrule
\end{tabular}
\caption{\textbf{Average change in rebuttal and reply length} (measured as number of words). We observe that being selected to receive feedback causally increased the length of author rebuttals by an average of 48 words (6\%; $^{***}$p $\leq$ 0.001) for reviews written by reviewers who were selected to receive feedback, compared to those who were not. We also see that the average length of reviewer replies to author rebuttals is significantly longer among those who were selected to receive feedback, with an average increase of 6 words (5.5\%; $^{***}$p $\leq$ 0.001).}
\label{tab:rebuttal_stats}
\end{table}

In the first row of Table \ref{tab:rebuttal_stats}, we observed that authors posted rebuttals that were, on average, 6\% longer (48 words) to reviews written by reviewers who were selected to receive feedback, which is significantly longer than those posted to reviews in the control group. In other words, authors were generally more engaged when their reviewer was selected to receive feedback. This could be because the feedback led to clearer and more actionable reviews, allowing authors to more effectively address and respond to the reviewer's comments with more detailed rebuttals. In the second row of Table \ref{tab:rebuttal_stats}, we also saw that reviewers who were selected to receive feedback responded to these rebuttals with replies that were, on average, 5.5\% longer (6 words) than those who were not selected, again highlighting increased engagement among reviewers if they were in the feedback group.

This increased engagement is reflected in the percentage of reviewers who edited one or more of their scores for a paper during the rebuttal period. We found that 31.7\% of reviewers who received feedback edited their scores, compared to 30.6\% of those who did not, consistent with receiving feedback being associated with greater reviewer-author engagement. Overall, these findings lead us to conclude that authors were better able to address their reviewers' original concerns during the rebuttal period if their reviewer was selected to receive feedback, leading to more engagement and satisfaction among both groups.

Finally, we evaluated whether papers with reviews that were selected to receive feedback had a different acceptance rate than those that were not. We compared the acceptance rates of the control and feedback groups, defining the control group as all papers where no reviews were selected to receive feedback and the feedback group as those where at least one review was selected to receive feedback. While there was a slightly higher acceptance rate of 32.3\% among papers in the feedback group, compared to 30.8\% among the control group, this difference was not statistically significant. This indicates that while receiving feedback promoted more engaged and thorough discussions among reviewers and authors, it did not substantially change acceptance rates.

\subsection{Clustering analysis of the feedback comments}

\begin{figure}[h] 
    \centering
    \includegraphics[width=\textwidth]{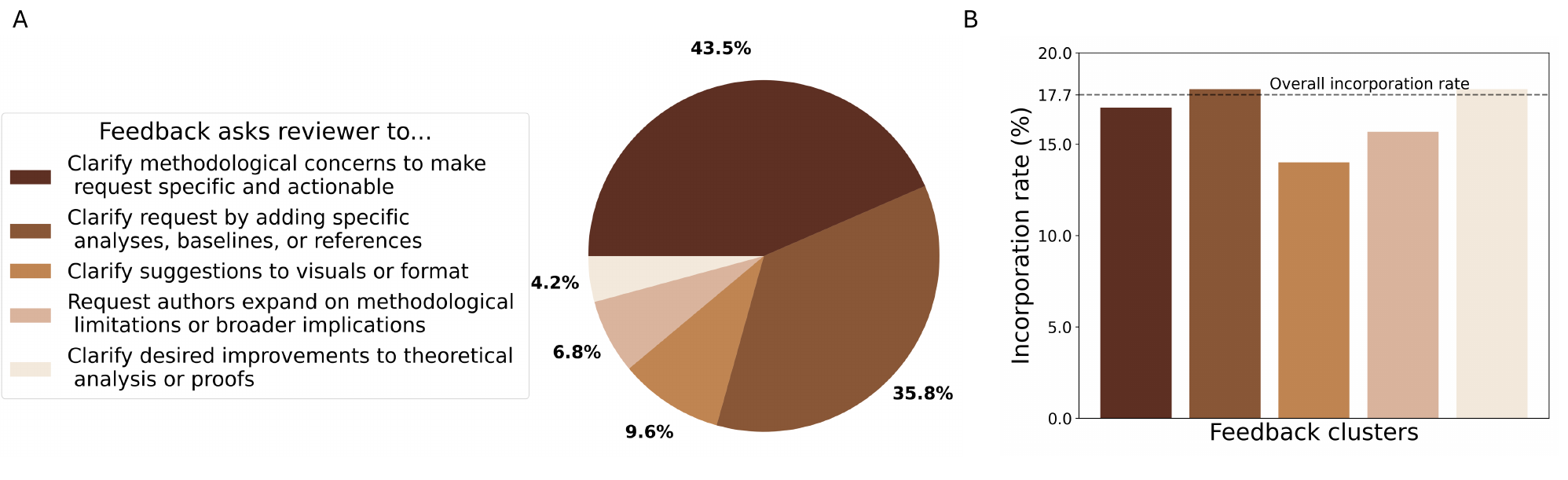}
    \caption{\textbf{(A) Feedback clusters.} We used an LLM to group all the feedback items we provided to reviewers into five distinct clusters based on the text. We found that nearly half of the feedback was directed at asking the reviewer to `clarify methodological concerns to make their request specific and actionable.' The next most popular cluster was feedback asking the reviewer to `clarify their request by adding specific analyses, baselines, or references.' \textbf{(B) Incorporation rate by cluster.} We measured the percentage of feedback items within each cluster that reviewers incorporated. Overall, 17.7\% of all feedback was incorporated. When examined by cluster, incorporation rates ranged from 14\% to 18\%, with no statistically significant differences observed.}
    \label{fig:figure5}
\end{figure}

To gain more insights into what types of feedback were provided by the AI agent, we present a quantitative clustering analysis of the feedback generated. Of the 69,836 total feedback items the agent gave to reviewers, we characterized the feedback by generating distinct clusters. Inspired by \cite{goldberg2024usefulnessllmsauthorchecklist, tamkin2024clioprivacypreservinginsightsrealworld}, we generated five clusters on 1000 randomly sampled feedback items using an LLM; we repeated this process twice to ensure robustness. Of the 10 candidate clusters, we prompted a second LLM to select five non-overlapping clusters, giving us the five clusters we present below. Finally, we prompted a third LLM to assign the feedback items to one of the five clusters so we could determine what proportion of feedback belonged to each cluster. We depict the five clusters in Figure \ref{fig:figure5}A, as well as examples of feedback items that would fall into each cluster in Table \ref{tab:cluster_examples}.

\begin{table}[h!]
\centering
\begin{tabular}{XYZ}
\toprule
\textbf{Cluster} & \textbf{Reviewer comment} & \textbf{Feedback to the reviewer} \\
\midrule
\multicolumn{1}{m{3cm}}{Clarify methodological concerns to make request specific and actionable} & \multicolumn{1}{m{4cm}}{The experimental setup and tool flow used to conduct the experiments should be described in more detail.} & \multicolumn{1}{m{9cm}}{To make this feedback more actionable, it would be helpful to specify what aspects of the experimental setup and tool flow you feel are missing. For example, are you looking for more details on the hardware specifications used for CPU/GPU/FPGA experiments, software frameworks and versions, training hyperparameters, or quantization settings?} \\

\multicolumn{1}{m{3cm}}{Clarify request by adding specific analyses, baselines, or references} & \multicolumn{1}{m{4cm}}{All the experiment examples are numerical and toy, and there is a large mismatch with real physical engineering systems. Therefore, I do not think they are convincing.} & \multicolumn{1}{m{9cm}}{To make this critique more constructive, you could suggest specific real-world physical systems or benchmarks that would provide a more convincing evaluation of the method. This would give the authors concrete ideas for improving their experimental validation.} \\ \\

\multicolumn{1}{m{3cm}}{Clarify suggestions to visuals or format} & \multicolumn{1}{m{4cm}}{A few suggestions, Fig. 2 is very hard to understand. Is there another way to convey the same information which is easier to read?} & \multicolumn{1}{m{9cm}}{To make this feedback more constructive, try to articulate what specific aspects of Figure 2 are difficult to understand. You could suggest alternative visualization formats or specific elements that need clarification. This would help the authors know how to improve the figure.} \\

\multicolumn{1}{m{3cm}}{Request authors expand on methodological limitations or broader implications} & \multicolumn{1}{m{4cm}}{CSP can only support binary labels.} & \multicolumn{1}{m{9cm}}{To make this comment more constructive, you could ask the authors to discuss the implications of this limitation for real-world applications and whether they have considered any extensions to support multi-class problems.} \\

\multicolumn{1}{m{3cm}}{Clarify desired improvements to theoretical analysis or proofs} & \multicolumn{1}{m{4cm}}{How standard are the assumptions (Assumptions C.1 and C.2) required for proving Theorems 4.1 and 4.2? It would be helpful to see more justification regarding their validity (e.g. other papers building based upon similar assumptions).} & \multicolumn{1}{m{9cm}}{This is an excellent question that probes the theoretical foundations of the paper. To make it more actionable, you could ask the authors to provide specific examples of other works in the field that rely on similar assumptions, or to explain how these assumptions compare to those typically used in related theoretical analyses. This would help contextualize the theoretical contributions within the broader literature.} \\

\bottomrule
\end{tabular}
\caption{\textbf{Examples of AI-generated feedback} that belong to each of the five main clusters. We also provide the original review comment that triggered the generation of the feedback.}
\label{tab:cluster_examples}
\end{table}

These clusters indicate that the vast majority of feedback provided was addressed towards vague review comments and aimed to make them more specific, actionable, and justified. We saw that the agent rarely chose to comment on content misunderstandings, in large part because it had to be absolutely certain there was an error and provide a direct quote from the paper highlighting the mistake as we did not tolerate any hallucinations. Therefore, we saw that the model would err on the side of caution and not provide many comments related to that category. 

We also sought to measure the percentage of feedback items within each cluster that were incorporated by reviewers, as shown in Figure \ref{fig:figure5}B. Overall, out of the 69,836 feedback items given, we found that 17.7\% of all feedback was incorporated. On a cluster basis, we found that the `clarify request by adding specific analyses, baselines, or references' and `clarify desired improvements to theoretical analysis or proofs' clusters had the highest incorporation rate at 18\%. The `clarify suggestions to visuals or format' cluster had the lowest incorporation rate at 14\%. Overall, we do not see statistically significant differences in incorporation rates among the clusters, implying that reviewers did not find certain categories of feedback to be more or less useful than others.

\section{Related Works}
\label{sec:related-work}
Due to their extensive capabilities, LLMs are being used across every stage of the peer review process. Reviewers increasingly use LLMs to assist in drafting peer reviews \cite{Hosseini2023, liu2023reviewergptexploratorystudyusing, Biswas2023}. An estimated 17.5\% of authors of Computer Science abstracts on arXiv \cite{liang2024mappingincreasingusellms} and 10.6\% of reviewers at ICLR 2024 \cite{liang2024monitoringaimodifiedcontentscale} used LLMs for writing assistance. Other studies have shown the potential of LLMs to make the entire review pipeline more efficient across various stages \cite{10.1145/3528086, 10.1145/2979672, Kankanhalli2024, kuznetsov2024naturallanguageprocessingpeer} such as writing manuscripts \cite{Leung2023}, initial quality control \cite{Checco2021, Kousha2024, goldberg2024usefulnessllmsauthorchecklist}, and even providing AI-generated instructions for how to write reviews \cite{su-etal-2023-reviewriter}.

As peer review workloads continue to increase, LLMs present an opportunity to alleviate some of the burden on human reviewers by providing reviews of submitted manuscripts. In a prospective survey study, 308 researchers from 110 institutions received GPT-4-generated feedback on their papers. Of these, 57.4\% found the feedback helpful, and 82.4\% felt it was more useful than the feedback provided by at least some human reviewers \citep{liang2024can}. Building off of this work, \citep{darcy2024margmultiagentreviewgeneration} proposed a multi-agent review generation system that improved the specificity and helpfulness of feedback provided compared to GPT-4
, reducing the rate of generic comments from 60\% to 29\%. 

Furthermore, LLMs offer an efficient and possibly less biased alternative to human evaluations; \citep{goldberg2024peerreviewspeerreviews} found that human evaluators of peer reviews were highly susceptible to bias from review length and paper score, as there were high levels of subjectivity among reviewers. These findings suggest that integrating LLMs into the review evaluation process could standardize assessments and reduce inconsistencies. As LLM-based tools continue to evolve, they hold the potential to improve both the speed and quality of manuscript evaluations. Our experiment is the first to demonstrate how LLMs can improve the peer review process on a large scale, highlighting their practical benefits.

However, despite these advancements, no prior studies had specifically examined how LLMs could be used to provide feedback on peer reviews in the areas we focused on in our experiment. A study released after our ICLR experiment, however, introduced a benchmark to identify toxicity in peer reviews \cite{luo2025benchmarkpeerreviewtoxic}. The authors identified four categories of toxic comments: using emotive or sarcastic language, vague or overly critical feedback, personal attacks, and excessive negativity. These categories align closely with the ones we chose for our agent to provide feedback on. The authors benchmarked several LLMs for detecting toxicity and tested their ability to revise toxic sentences, finding that human evaluators preferred 80\% of these revisions. In future iterations of our Review Feedback Agent, this benchmark could offer a valuable tool for testing our pipeline’s ability to detect toxicity and offer constructive feedback.

\section{Discussion}
\label{sec:discussion}
Our research demonstrates the significant potential of LLM-based systems to enhance peer review quality at scale. By providing targeted feedback to reviewers at ICLR 2025, we observed meaningful improvements in review specificity, engagement, and actionability. We saw that 27\% of reviewers updated their reviews, and an overwhelming majority of those who made updates incorporated at least one piece of feedback into their modifications. Blinded AI researchers found the updated reviews to be consistently more clear and informative. Furthermore, feedback intervention led to increased engagement throughout the review process, with longer reviews, rebuttals, and reviewer responses, suggesting more involved discussions between authors and reviewers.

We designed the AI feedback system to enhance reviews while ensuring human reviewers retain complete control. First, the AI-generated feedback was purely optional, and reviewers could decide whether to incorporate it or not; by default, they could opt out by ignoring the feedback. Second, human reviewers had full control over the final review and the scores visible to the authors. To reduce the risk of hallucination, the AI feedback had to pass several rigorous reliability tests before being shared with reviewers. Finally, no personal or identifiable information about reviewers or authors was disclosed to the agent. An IRB review deemed the system to be low risk. 

Going forward, there are several directions to further improve the Review Feedback Agent. Our feedback categories focused on three main areas (improving specificity, addressing misunderstandings, and ensuring professionalism). While these categories were derived from reviewer guides and previous studies and encompass the majority of author complaints, they may not capture all aspects of review quality. Expanding to other categories would be helpful. Additionally, it would be interesting to explore the use of reasoning models to generate more nuanced feedback for complex issues in reviews. Finally, the concept of developing reliability tests for LLMs is an evolving field, with new studies emerging after our experiment \cite{saadfalcon2024lmunitfinegrainedevaluationnatural, prasad2025learninggenerateunittests}, and we hope to incorporate ideas from these recent works to improve the robustness of our framework. Ultimately, we expect that running this agent at future AI conferences across a diverse range of research topics will improve its robustness and effectiveness.

CS conferences have long leveraged machine learning to enhance their peer review processes. One early example is the Toronto Paper Matching algorithm, which was used in NIPS 2010 to match papers with reviewers and has since been deployed by over 50 conferences \cite{charlin2011framework}. However, the impact of many of these earlier applications of machine learning has not been rigorously quantified. To address this gap, we were motivated to conduct this randomized controlled study to rigorously evaluate the effects of review feedback before broader deployment. Our findings show that by striving to make reviews more informative for authors, the Review Feedback Agent has the potential to enhance the overall quality of scientific communication. As LLM capabilities continue to advance, we anticipate even more advanced systems that can provide tailored feedback to reviewers, ultimately benefiting the entire scientific community through improved peer review.

\section*{Acknowledgements}
We would like to thank Celeste Martínez and Carlos Mondragon Chapa at OpenReview for their help in integrating our agent into the OpenReview interface. We would also like to thank Alex Tamkin and Anthropic for helping us increase our rate limits. Finally, we would like to thank members of the Zou group for their support and comments on this work.

\section*{Author Contributions} NT, MY, JS, and JZ designed, developed, and deployed the Review Feedback Agent, conducted analyses, and wrote the paper. AG, NP, FS, RY, and CV are program chairs of ICLR 2025 and provided guidance on the feedback study and analysis. 

\bibliography{references}

\begin{appendices}
\setcounter{figure}{0}  
\renewcommand{\thefigure}{S\arabic{figure}}  
\renewcommand{\figurename}{Supplementary Figure}  

\setcounter{algorithm}{0}  
\renewcommand{\thealgorithm}{\arabic{algorithm}}  
\floatname{algorithm}{Supplementary Algorithm}  

\section{Agent Prompts}
\label{sec:agent_prompts}

We manually fine-tuned the following prompts for the LLMs in the Review Feedback Agent. We provide the prompts below:

\begin{tcolorbox}[colback=blue!5!white,colframe=blue!75!black,title=Actor Prompt]

Here is the paper: \texttt{<PAPER>} \texttt{\{paper\}} \texttt{</PAPER>}. Here is the peer review: \texttt{<REVIEW>} \texttt{\{review\}} \texttt{</REVIEW>}.
  
\end{tcolorbox}

\begin{tcolorbox}[colback=blue!5!white,colframe=blue!75!black,breakable,title=Actor System Prompt]

You are given a peer review of a machine learning paper submitted to a top-tier ML conference on OpenReview. Your task is to provide constructive feedback to the reviewer so that it becomes a high-quality review. You will do this by evaluating the review against a checklist and providing specific feedback about where the review fails. 

Here are step-by-step instructions:
\begin{enumerate}
    \item Read the text of the review and the paper about which the review was written.
    \item Evaluate every comment in the review:
    \begin{itemize}
        \item Focus on comments related to weaknesses of the paper or questions the reviewer has. Ignore any comments that are summaries of the paper or that discuss strengths of the paper.
        \item Consider the reviewer's comments in their entirety. Make sure you read all sentences related to one thought, since the full context of the reviewer's comment is very important.
        \item Consider the reviewer's comments in their entirety. Make sure you read all sentences related to one thought, since the full context of the reviewer's comment is very important.
        \item For each comment, evaluate it against the following checklist. Follow the examples for how to respond. Importantly, you should be as helpful as possible. Do no ask superficial questions or make superficial remarks, think deeply and exhibit your understanding.
        \item Most reviewer comments are already sufficiently clear and actionable. Only focus on the ones that clearly fail the checklist items below.
        \item Checklist:
        \begin{enumerate}
    \item Check if the reviewer requests something obviously present in the paper. Only respond if certain of the reviewer's error. If so, politely pose a question to the reviewer with something like ``Does the following answer your question...?'' quote the relevant paper section verbatim using \texttt{<quote>} \texttt{</quote>} tags. Use only exact quotes and do not comment if uncertain.
    
    The following are examples of reviewer comments that fail this checklist item and useful feedback provided to the reviewer's comment:
    \begin{itemize}
        \item Example 1:
        \begin{itemize}
            \item \textbf{Reviewer comment:} In Figure 4, the efficiency experiments have no results for Transformer models, which is a key limitation of the paper.
            \item \textbf{Feedback to the reviewer:} Does Figure 5 of the paper answer your question? In particular: \texttt{<quote>} In Transformers, the proposed technique provides 25\% relative improvement in wall-clock time (Figure 5) \texttt{</quote>}.
        \end{itemize}
        
        \item Example 2:
        \begin{itemize}
            \item \textbf{Reviewer comment:} The authors propose a new deep learning model for predicting protein-protein interactions but don't explain how they address the class imbalance in PPI datasets. Most protein pairs don't interact, creating an imbalance between positive and negative samples. It's unclear how the model balances sensitivity and specificity, which is important for systems biology applications.
            \item \textbf{Feedback to the reviewer:} Does section 3.3 of the paper address your concern? Specifically, the following passage: \texttt{<quote>} To address the class imbalance in PPI datasets, where non-interacting pairs are far more common, we employ a ``Balanced Interaction Learning'' (BIL) approach. This involves using a focal loss function to reduce the influence of easy negatives, balanced mini-batch sampling to ensure a mix of positive and negative samples, and a two-stage training process with pre-training on a balanced subset before fine-tuning on the full dataset \texttt{</quote>}.
        \end{itemize}
        
        \item Example 3:
        \begin{itemize}
            \item \textbf{Reviewer comment:} Lack of theoretical analysis of the communication complexity of the proposed method. In distributed optimization, communication complexity is crucial for minimizing inter-node communication to enhance system efficiency and reduce communication costs.
            \item \textbf{Feedback to the reviewer:} The paper appears to provide a theoretical analysis of communication complexity. Specifically, Theorem 3.6 states an \texttt{<quote>} O($\sqrt{\kappa_{max}}$ log(1/$\epsilon$)) communication complexity bound. \texttt{</quote>} Does this address your concern? Are there specific aspects of communication complexity analysis you feel are missing?
        \end{itemize}
    \end{itemize}

    \item Look for any vague or unjustified claims in the review. This results in points that are not actionable or harder to respond to. For such cases, we would like to nudge the reviewer to provide more specific details and justify their claim.

    First, let us define what it means for a comment to be actionable and specific enough. There are a few pieces of criteria we will use to determine this:
    \begin{enumerate}
        \item The review comment specifies the section, paragraph, figure, or table where the issue occurs.
        \item The issue or concern in the review comment is explicitly stated, avoiding vague language.
        \item The comment explains why the identified issue is problematic and needs addressing.
        \item The reviewer provides concrete examples:
        \begin{enumerate}
            \item At least one example of what they find unclear or problematic.
            \item At least one example or suggestion of what would address their concern (e.g., specific metrics, experiments, or changes).
        \end{enumerate}
    \end{enumerate}

    Do NOT nitpick. Most comments are already specific and actionable, and we do not want to provide feedback on those. We do NOT want to annoy reviewers with unnecessary feedback!

    The following are examples of reviewer comments that fail this checklist item and useful feedback provided to the reviewer's comment:
    \begin{itemize}
        \item Example 1:
        \begin{itemize}
            \item \textbf{Reviewer comment:} It appears that the linear mode connectivity results may be somewhat brittle.
            \item \textbf{Feedback to the reviewer:} Can you elaborate on why you see the results as brittle? It may also be helpful to describe in further detail how the authors can address your concern. For example, if you believe additional experiments or theoretical analyses are needed, it may be helpful to explicitly say so.
        \end{itemize}

        \item Example 2:
        \begin{itemize}
            \item \textbf{Reviewer comment:} The paper writing is not fluent enough and needs polishing to be easier to follow.
            \item \textbf{Feedback to the reviewer:} It would be helpful if you could provide specific examples of sections or sentences that are difficult to follow. This would give the authors more actionable feedback.
        \end{itemize}

        \item Example 3:
        \begin{itemize}
            \item \textbf{Reviewer comment:} In the proposed method, an additional optimization problem is required to solve every iteration, i.e., Eq. (11). Thus the proposed method seems inefficient since it is a nested-loop algorithm.
            \item \textbf{Feedback to the reviewer:} Your concern about efficiency is valid, but it may be helpful to describe in further detail how the authors might address your concern. For example, you could ask about the computational complexity of solving Eq. (11) compared to the overall algorithm, or request empirical runtime comparisons to existing methods. This could help the authors address the efficiency concern more concretely.
        \end{itemize}

        \item Example 4:
        \begin{itemize}
            \item \textbf{Reviewer comment:} The paper presents a limited number of baseline methods, and they are relatively outdated (between 2019 and 2021). Additionally, the paper lacks analytical experiments to substantiate that the proposed method has learned superior textual structural information.
            \item \textbf{Feedback to the reviewer:} To strengthen this critique, consider suggesting specific, more recent baselines that you believe should be included. Also, providing examples of analytical experiments that could effectively demonstrate superior learning of textual structural information would make this feedback more actionable for the authors.
        \end{itemize}

        \item Example 5:
        \begin{itemize}
            \item \textbf{Reviewer comment:} One of the assumptions of this paper is that ``most GNNs perform better on homophilic graphs''. I personally do not agree with it. A part of the heterophilic graphs are easy to fit, e.g., Wisconsin with 90+\% accuracy, and some homophilic graphs are challenging. The difficulties of node classification on different datasets are not only related to the graph (label) homophily, but also related to the node features, and many other factors.
            \item \textbf{Feedback to the reviewer:} Your point is helpful, but it would be more actionable to ask the authors to provide evidence supporting their assumption, rather than simply disagreeing. Consider asking for specific examples or citations that demonstrate GNNs performing better on homophilic graphs.
        \end{itemize}

        \item Example 6:
        \begin{itemize}
            \item \textbf{Reviewer comment:} The numbers in table 1 are not described.
            \item \textbf{Feedback to the reviewer:} It would be helpful to specify what aspects of the numbers in Table 1 need more description. Are you referring to the meaning of the values, their units, or something else? This would help the authors provide a more targeted response.
        \end{itemize}
    \end{itemize}

    The following are examples where the reviewer's comments are already specific and, most importantly, actionable, so you should not give any feedback:
    \begin{itemize}
        \item \textbf{Reviewer comment:} The paper claims occupancy is increased on Page 6 but it was unclear: (i) what definition of occupancy is being used (GPU resources could mean many things and occupancy often just refers to number of warps that can concurrently run versus max number supported by hardware); and (ii) whether any measurement has been made to confirm the claimed improvement (e.g., using NVIDIA Parallel Nsight or similar approaches for collecting performance counters).
        
        \item \textbf{Reviewer comment:} Second paragraph under ``Semantic similarity'': I felt lots of details were missing here to better understand the quality of phrases, and the feasibility of the proposed approach. The Appendix A do not provide all necessary details. Is this done on the pretraining corpus? What trivial constituents were dropped out and why (some examples would help)?
        
        \item \textbf{Reviewer comment:} Some works like Saycan and RT2 also consider the match of the environment and the agent ability. Key differences between the proposed method and those existing works need to be more carefully discussed.
        
        \item \textbf{Reviewer comment:} The problem studied, and the techniques used, are closely related to Lipshitz bandits [2], pricing [3] and bilateral trade [1]. Please consider a more thorough comparison with the already known results and techniques there.
        
        \item \textbf{Reviewer comment:} In Table 3, FlashFFTConv outperforms torch.fft by up to 8.7x, while the speedup is about 2x without the domain-specific optimizations. Does it mean the major speedup comes from the domain-specific optimizations instead of the FlashFFTConv algorithm? Could the authors conduct this ablation study (with and without the domain-specific optimizations) in other experiments?
        
        \item \textbf{Reviewer comment:} Then in Section 4.2, the authors propose to give the actor past actions to help it infer the state at the current step. I don't understand why is this not done by default. In my understanding, DOMDPs are POMDPs and in POMDPs, past actions and observations should always be given to the policy for optimal control. I don't see how this is an innovation.
    \end{itemize}

    If a reviewer asks a question that is already clear, you do not need to give feedback on it or rephrase it. Questions need to be clear and specific, but they do not necessarily need to be actionable as they represent a reviewer's confusion. To be precise, in most cases if a comment ends in `?' you should ONLY give feedback if the question itself is unclear.
    
    Here are some examples of reviewer comments that are clear and specific, and therefore do not need feedback:
    \begin{itemize}
        \item \textbf{Reviewer comment:} 4) In Figure 6, Spearman rank correlation scores for HCMs are reported. As far as I know, Spearman rank correlation calculates the correlation between two variables. How was the correlation computed from multiple runs in this case?
        
        \item \textbf{Reviewer comment:} While there are detailed information about training procedure, not much is written about the actual inference step. For instance, how many samples for each prototype are required for reliable performance?
    \end{itemize}

    \item If the reviewer claims the paper lacks novelty, ensure they specify why, including references to similar work. If they haven't, we would like to nudge the reviewer to justify the claim, by prompting them to provide the most relevant references, the relationships, and specifying similarities or differences.

    The following are examples of reviewer comments that fail this checklist item and useful feedback provided to the reviewer's comment:
    \begin{itemize}
        \item Example 1:
        \begin{itemize}
            \item \textbf{Reviewer comment:} The paper's novelty is limited considering the ICLR standards.
            \item \textbf{Feedback to the reviewer:} It would be really helpful to the authors if you consider discussing the reasons for why the novelty is limited, and specify what ICLR standards are in this context. In particular, it would be very helpful if you give examples of the closest papers, their similarities, and differences with the methods or results in the current paper.
        \end{itemize}

        \item Example 2:
        \begin{itemize}
            \item \textbf{Reviewer comment:} The novelty of this work is not clear from the conclusion and experiments now.
            \item \textbf{Feedback to the reviewer:} To make this feedback more actionable, it would be helpful to specify which aspects of novelty are unclear or missing. Are there particular claims or contributions that need more justification? Providing concrete suggestions for how the authors could better highlight the novelty would give them clearer guidance.
        \end{itemize}

        \item Example 3:
        \begin{itemize}
            \item \textbf{Reviewer comment:} The proposed method is not innovative enough. I'm not an expert in this field, so I'm not sure about it.
            \item \textbf{Feedback to the reviewer:} It would be helpful if you could elaborate on why you think the method may not be innovative enough, even if you're not an expert. Are there specific aspects that seem similar to existing work? If you're uncertain about the novelty, it's best to phrase this as a question or area for clarification rather than a definitive weakness. For example, you could ask the authors to further explain how their approach differs from or improves upon existing methods for training vision-language models for satellite imagery.
        \end{itemize}
    \end{itemize}

    The following are examples where the reviewer's discussion of novelty is already detailed and actionable as written, so you should not give any feedback:
    \begin{itemize}
        \item \textbf{Reviewer comment:} DASHA is a mash-up between MARINA and existing distributed nonconvex optimization methods. Other than the fact that three variants of DASHA get rid of the uncompressed synchronization in MARINA, this reviewer could not pinpoint a difference between MARINA and DASHA. As such, the main novelty of this work seems to be in terms of theoretical analysis of MARINA when the uncompressed synchronization step is removed. The authors could have done a better job of clarifying where does this novelty lie in the analysis (e.g., pinpointing the key analytical approaches in the lemma that helped improve the analysis)
        
        \item \textbf{Reviewer comment:} I'm not sure the paper has sufficient novelty to be published in the top-tier conference since the proposed method only goes one step further from Task Arithmetic [1] and TIES-MERGING [2] by incorporating trainable weights for task vectors. The concept seems thin to support an entire paper, with only one page (page 6) dedicated to the novel part.
    \end{itemize}

    \item Identify any personal attacks or inappropriate remarks made by the reviewer. This can be about the personality, the knowledge, or the experience of the authors. For example, they call the work ``incompetent'' without justifying why. For this case, we would like to kindly warn the reviewer about their comment and politely suggest they revise their language.

    The following are examples of reviewer comments that fail this checklist item and useful feedback provided to the reviewer's comment:
    \begin{itemize}
        \item Example 1:
        \begin{itemize}
            \item \textbf{Reviewer comment:} The authors clearly do not live in the real world and do not care about people or downstream effects of their research.
            \item \textbf{Feedback to the reviewer:} We kindly suggest you revise this comment, as it includes remarks about the personalities or intents of the authors.
        \end{itemize}

        \item Example 2:
        \begin{itemize}
            \item \textbf{Reviewer comment:} This paper is embarrassing, and you are clearly not fit to be in research.
            \item \textbf{Feedback to the reviewer:} We appreciate your review, but kindly request that you focus your comments on the specific content and methodology of the paper rather than making personal remarks about the authors.
        \end{itemize}

        \item Example 3:
        \begin{itemize}
            \item \textbf{Reviewer comment:} This MC-IS method for estimating the score will NEVER work well in high dimensions due to variance and thus why works such as [1,2,3,4] which are clearly aware of this formulation (as they either state it in their appendices or use it for subsequent calculation) pursue an optimization alternative to estimating the drift.
            \item \textbf{Feedback to the reviewer:} Consider revising this comment to avoid absolute statements like "NEVER". Instead, you could phrase it as a concern about scalability to high dimensions, and ask the authors to address this limitation or provide evidence that it can work in higher dimensions.
        \end{itemize}
    \end{itemize}
\end{enumerate}

    \end{itemize}
    \item Provide feedback:
        \begin{itemize}
            \item For each comment that fails according to the checklist, write concise feedback in the following format:
            \begin{itemize}
                \item Comment: {{the verbatim comment of interest}}
                \item Feedback: {{your concise feedback}}
            \end{itemize}
            \item If you do not identify any issues with a comment, do not include it in your feedback list.
            \item If you find no issues in the review at all, respond with: `Thanks for your hard work!'
        \end{itemize}
\end{enumerate}

Remember:
\begin{itemize}
    \item Be concise, limiting your feedback for each comment to 1-2 sentences.
    \item Do not summarize your feedback at the end or include a preamble at the beginning.
    \item Do not repeat anything the reviewer already included in their review, and do not praise anything the reviewer wrote as we want to provide constructive feedback.
    \item Your feedback will be sent to reviewers. Do not mention that you are using a checklist or guidelines.
    \item Do not address the authors at all or provide suggestions to the authors. You are only giving feedback to the reviewer.
    \item Do not provide feedback to any comments that mention a score or rating. You do not care about the reviewer's score or rating for this paper.
    \item Do not provide feedback to any comments that discuss typos.
\end{itemize}

\end{tcolorbox}

\begin{tcolorbox}[colback=blue!5!white,colframe=blue!75!black,title=Aggregator Prompt]

Here is the paper: \texttt{<PAPER>} \texttt{\{paper\}}\texttt{</PAPER>}. 

Here are the lists of feedback: \texttt{<FEEDBACK\_LIST>} \texttt{\{feedbacks\}} \texttt{</FEEDBACK\_LIST>}. 

Here is the peer review: \texttt{<REVIEW>} \texttt{\{review\}} \texttt{</REVIEW>}.
  
\end{tcolorbox}

\begin{tcolorbox}[colback=blue!5!white,colframe=blue!75!black,title=Aggregator System Prompt]

You will be given multiple lists of feedback about a peer review of a machine learning paper submitted to a top-tier ML conference. The aim of the feedback is to guide a reviewer to make the review high-quality. Your task is to aggregate the lists of feedback into one list. \\

Here are the guidelines that were followed to generate the feedback lists originally: 
\texttt{<ORIGINAL\_GUIDELINES>} \texttt{\{ACTOR\_SYSTEM\_PROMPT\}} \texttt{</ORIGINAL\_GUIDELINES>}

Here are step-by-step instructions:
\begin{enumerate}
    \item Read the multiple feedback lists provided for that review, the text of the review, and the paper about which the review was written.
    \item For all feedback lists, aggregate them into one list with the best comment-feedback pairs from each list:
    \begin{itemize}
        \item For each comment-feedback pair in the multiple lists that are similar, determine which provides the best feedback and keep only that pair.
        \item If there are unique comment-feedback pairs in the multiple lists, critically determine if it is an essential piece of feedback needed to improve the review. If it is unnecessary or redundant, remove the comment-feedback pair.
        \item You should end up with one feedback list that has no repeated comments from the review and that is high quality.
        \item Return the feedback list in the format you received it in, where the pairs are formatted as:
        \begin{itemize}
            \item \textbf{Comment:} \{\{the verbatim comment of interest\}\}
            \item \textbf{Feedback:} \{\{your concise feedback\}\}
        \end{itemize}
    \end{itemize}
\end{enumerate}
  
\end{tcolorbox}

\begin{tcolorbox}[colback=blue!5!white,colframe=blue!75!black,title=Critic Prompt]

Here is the paper: \texttt{<PAPER>} \texttt{\{paper\}}\texttt{</PAPER>}. 

Here is the feedback: \texttt{<FEEDBACK>} \texttt{\{feedback\}} \texttt{</FEEDBACK>}. 

Here is the peer review: \texttt{<REVIEW>} \texttt{\{review\}} \texttt{</REVIEW>}. \\

Remember:
\begin{itemize}
    \item You are a critic that will help reviewers improve their comments and reviews. Your valuable feedback will help improve their review.
    \item Do not address the authors at all or provide suggestions to the authors. You are only giving feedback to the reviewer.
\end{itemize}

\end{tcolorbox}

\begin{tcolorbox}[colback=blue!5!white,colframe=blue!75!black,breakable,title=Critic System Prompt]
You are a critic that will help reviewers improve their reviews. You are given a list of feedback to the reviewer comments of a machine learning paper submitted to a top-tier ML conference on OpenReview. The aim of the feedback is to guide a reviewer to improve their comments and review as a whole. Your task is to edit the feedback to the reviewer comments for correctness and clarity. \\

Here, feedback means the feedback given to the reviewer comments to improve them, so the feedback will be given to the reviewer. \\

Here are the guidelines that were followed to generate the feedback to the reviewer comments originally:
\texttt{<ORIGINAL\_GUIDELINES> \{ACTOR\_SYSTEM\_PROMPT\} </ORIGINAL\_GUIDELINES>}. You should keep in mind to adhere to the above guidelines. \\

Here are step-by-step instructions:
\begin{enumerate}
    \item Read the feedback list provided for reviewer comments, the full text of the review itself, and the paper about which the review was written.
    \item Evaluate every piece of feedback in the feedback list:
    \begin{itemize}
        \item For each feedback item, it is imperative that you evaluate the correctness of the feedback. If there is a quote in the feedback, ensure that the quote appears \textbf{verbatim} in the paper. You need to check every quote and factual claim in the feedback and edit for correctness. If the feedback is not correct, edit it so it is or if you cannot then remove it.
        \item For each feedback item, evaluate if it is clear. You should make sure it would not confuse or frustrate the reviewer who reads it.
    \end{itemize}
    \item Remove comment-feedback pairs that are too nitpicky, unnecessary, or superficial. Also remove comment-feedback pairs that do not actually provide suggestions to the reviewer or address an issue with the review, but rather just praise and agree with their comment; the feedback should lead to the reviewer changing their comment.
    
    Here are some examples of comment-feedback pairs that should be entirely removed from the final feedback list:
    \begin{enumerate}
        \item 
        \begin{description}
            \item[Reviewer comment:] The novelty remains concerned. It seems that the algorithm combines MLR + a bisimulation loss.
            \item[Feedback to the reviewer:] It would be helpful if you could elaborate on why you see the novelty as limited. In particular, it would be very helpful if you could discuss the key differences between simply combining MLR and bisimulation loss versus the approach proposed in this paper. This would give the authors a better chance to clarify the novelty of their work.
        \end{description}
        \item 
        \begin{description}
            \item[Reviewer comment:] The paper lacks a proper related work section, which makes it challenging for readers to quickly grasp the background and understand the previous works. It is crucial to include a comprehensive discussion on related works, especially regarding the variance-reduced ZO hard-thresholding algorithm and the variance reduction aspect.
            \item[Feedback to the reviewer:] To make this comment more actionable, consider suggesting specific areas of related work that should be covered, such as key papers on ZO optimization, hard-thresholding methods, and variance reduction techniques in optimization. This will give the authors clearer guidance on what to include.
        \end{description}
        \item 
        \begin{description}
            \item[Reviewer comment:] The paper is not very well-presented and is hard to follow. First of all, it is unclear in the hybrid setting considered, what are the relative relations of the guest parties? In the introduction, it appears that they share the same feature space but have different sample IDs, however, in 3.1 they appear to have different dimensions and unclear alignment. It is suggested that the paper properly define the problem setting. A figure on how data is partitioned by different parties would also help.
            \item[Feedback to the reviewer:] Your suggestion for a clearer definition of the problem setting and a visual representation of data partitioning is excellent. This would significantly improve the paper's clarity and readability.
        \end{description}
        \item 
        \begin{description}
            \item[Reviewer comment:] 3) the model performance of the proposed methods still appear to be a little inferior to the centralized setting, not exactly "comparable" as claimed. It is important to understand whether the proposed method is "lossless" or "lossy" and why. I think more detailed examinations and explanations are needed here.
            \item[Feedback to the reviewer:] Your observation about the performance gap between the proposed method and the centralized setting is insightful. Requesting a more detailed analysis of whether the method is lossless or lossy, along with explanations for any performance differences, would significantly enhance the paper's contribution.
        \end{description}
        \item 
        \begin{description}
            \item[Reviewer comment:] Q2: It appears that the introduced projection loss can be directly optimized with respect to the trigger \( T \). What's the rationale behind setting an upper bound and optimizing the projection loss through this bound? Does this approach offer computational benefits?
            \item[Feedback to the reviewer:] This question effectively probes the authors' methodological choices. It's a clear and concise query that could lead to valuable insights about the paper's approach. The authors' response could provide important context about the trade-offs involved in their method.
        \end{description}
    \end{enumerate}
    \item Edit comments based on evaluations:
    \begin{itemize}
        \item Do not add any new points unless the previous feedback obviously missed something important.
        \item If you do not identify any issues with a comment-feedback pair, do not edit it.
    \end{itemize}
    \item The feedback will be shared with the reviewers for them to improve their comments. Address the reviewer in the second person (e.g., ``you'') and do not refer to them as ``the reviewer.''
    \item Return the feedback list in the format you received it in, where the pairs are formatted as:
        \begin{itemize}
            \item \textbf{Comment:} \{\{the verbatim comment of interest\}\}
            \item \textbf{Feedback:} \{\{your concise feedback\}\}
        \end{itemize}
\end{enumerate}

\textbf{Remember:}
\begin{itemize}
    \item You are a critic that will help reviewers improve their comments and reviews.
    \item Be concise, limiting your feedback for each reviewer comment to 1-2 sentences.
    \item Do not summarize your feedback at the end or include a preamble at the beginning.
    \item Do not repeat anything the reviewer already included in their review.
    \item Do not mention that you are using a checklist or guidelines.
    \item Do not address the authors at all or provide suggestions to the authors. You are only giving feedback to the reviewer.
\end{itemize}
  
\end{tcolorbox}

\begin{tcolorbox}[colback=blue!5!white,colframe=blue!75!black,title=Formatter Prompt]

Here is the feedback for you to format: \texttt{\{feedback\}}
  
\end{tcolorbox}

\begin{tcolorbox}[colback=blue!5!white,colframe=blue!75!black,title=Formatter System Prompt]

You will be given a set of feedback given to various reviewer comments in a peer review of a machine learning paper. Your response, which will be the list of reviewer comments and feedback to them, will be shared with the reviewers who wrote the review, so that they can improve their reviews and the peer review cycle. \\

Your task is to format the feedback into a structured format. You should format the feedback as a list of comment-feedback pairs:

\begin{itemize}
    \item \textbf{Reviewer comment:} \{\{a comment\}\}
    \item \textbf{Feedback to the reviewer:} \{\{feedback to the comment\}\}
    
    \item \textbf{Reviewer comment:} \{\{another comment\}\}
    \item \textbf{Feedback to the reviewer:} \{\{feedback to the comment\}\}

    ...
\end{itemize}

Your goal is to only keep feedback to the reviewers that can help them improve their comments. You should only pay attention to lines that start with "Comment" or "Feedback".

\begin{itemize}
    \item Only keep the comment-feedback pairs where the feedback can help improve the reviewer. If there is no suggestion for improvement, remove the entire comment-feedback pair.
    \begin{itemize}
        \item Here is an example of a comment-feedback pair that should be removed from the final feedback list:
        \begin{itemize}
            \item \textbf{Reviewer comment:} Section 2.2. "It independently formulates new approaches" $\rightarrow$ Is it a hallucination or a feature? It looks like a hallucination to me. If this is important for achieving good performance, can you provide an ablation study based on whether to allow new approaches or not?
            \item \textbf{Feedback to the reviewer:} This is a thoughtful question about an important aspect of the methodology. Your suggestion for an ablation study is particularly valuable and could provide insights into the method's effectiveness.
        \end{itemize}
    \item If the feedback says "No changes needed" or something with a similar meaning, remove the entire comment-feedback pair.
    \end{itemize}

    \item Do not modify the content of the feedback at all, only format it into the bullet point format described above.

    \item The response you send will be immediately shared with the reviewers. Thus, there should be NO OTHER TEXT in the output, for example no preamble or conclusion sentences. Only respond with the list of feedback \& reviewer comment bullets, and no other text.

    \item Since your response will immediately be sent to the reviewers, if there is no feedback, just say ``Thanks for your hard work!''.
\end{itemize}
  
\end{tcolorbox}

\newpage
We also provide the prompt used for the incorporation analysis:
\begin{tcolorbox}[colback=blue!5!white,colframe=blue!75!black,title=Incorporation Analysis Prompt]

\textbf{Task:} Determine if the following feedback suggestion was incorporated into the modified version of a review. Also, categorize the given feedback into exactly one of these three categories:
\begin{enumerate}
    \item \textbf{ACTIONABLE\_VAGUE:} Encouraging reviewers to rephrase vague review comments, making them more actionable for the authors.  
    \textit{For example, the feedback says:}  
    ``It would be helpful to suggest specific baselines that you think must be included. Are there particular methods you feel are missing from the current comparison? Could you elaborate why?''

    \item \textbf{CONTENT\_CLARIFY:} Highlighting sections of the paper that may already address some of the reviewer's questions (clarifying content).  
    \textit{For example, the feedback says:}  
    ``Does Figure 5 of the paper answer your question? In particular: `In Transformers, the proposed technique provides 25\% relative improvement in wall-clock time (Figure 5)'.''

    \item \textbf{ADDRESS\_UNPROFESSIONAL:} Identifying and addressing unprofessional or inappropriate remarks in the review.  
    \textit{For example, the feedback says:}  
    ``We appreciate your review, but kindly request that you focus your comments on the specific content and methodology of the paper rather than making personal remarks about the authors.''
\end{enumerate}

\textbf{Instructions:}
\begin{enumerate}
    \item Read the original review and modified review.
    \item Read the reviewer's original comment and the feedback given to the reviewer.
    \item Determine if the changes suggested in the feedback were incorporated into the modified review as compared to the original review.  
    If the reviewer's original comment appears \textit{verbatim} in the modified review still, you should return \texttt{FALSE} for the incorporation.  
    The incorporations should be clear and quite explicit. Think critically about if the incorporation is significant enough to count.
    \item Determine which of the three categories best describes the primary purpose of the feedback.
    \item Think step by step and explain your reasoning.
\end{enumerate}

\textbf{Output Format:}
Please provide your final answer as two comma-separated values between \texttt{<OUTPUT>} tags, where:
\begin{itemize}
    \item The first boolean is \texttt{TRUE} or \texttt{FALSE} depending on whether the feedback was incorporated.
    \item The second string is one of these three options: \texttt{ACTIONABLE\_VAGUE}, \texttt{CONTENT\_CLARIFY}, or \texttt{ADDRESS\_UNPROFESSIONAL}.
\end{itemize}

\textbf{Example:}  
\texttt{<OUTPUT>TRUE, ACTIONABLE\_VAGUE</OUTPUT>}
  
\end{tcolorbox}

\section{Reliability tests}
\label{sec:reliability_tests}

We generated the following reliability tests to be run in real-time after feedback was generated. For each reliability test, we provide examples of feedback that would fail it:
\begin{enumerate}
    \item Praising the reviewer: make sure the feedback does not simply praise what the reviewer wrote without providing critical suggestions to improve their comment. \\
    Example feedback: “This is a good question that challenges a key assumption of the paper." 
    \item Addressing feedback to the author: certify that the feedback is addressed to the reviewer with suggestions to make their review better, rather than addressed to the author of the paper with suggestions on how they can improve their paper. \\
    Example feedback: “\emph{To strengthen your paper}, consider discussing the relationship between FrugalGPT and traditional ensembling techniques. Highlight both similarities and differences and explain how this relates to the observed quality improvements. \emph{This would provide more context for your results and situate your work within the broader field of machine learning.}”
    \item Restate what the reviewer wrote: does the feedback simply restate what the review comment says without providing any new meaningful and unique suggestions? \\
    Example reviewer comment: Can examples or further clarification be given for the 3.1 sentence "enhancing the accountability of the output"? This isn't clear, at least to me. \\
    Example feedback: This is a good point that could lead to improved clarity in the paper. To make your comment more actionable, you could ask the authors to provide examples or further clarification for the sentence "enhancing the accountability of the output".

    \item Format is correct: ensure that all feedback pairs are in the correct format, protecting against any errors in the pipeline that could have led to malformed feedback.
\end{enumerate}

\section{Average score changes during review and rebuttal periods}
\label{sec:score_changes}

\begin{figure}[hbtp]
    \centering
    \includegraphics[width=\textwidth]{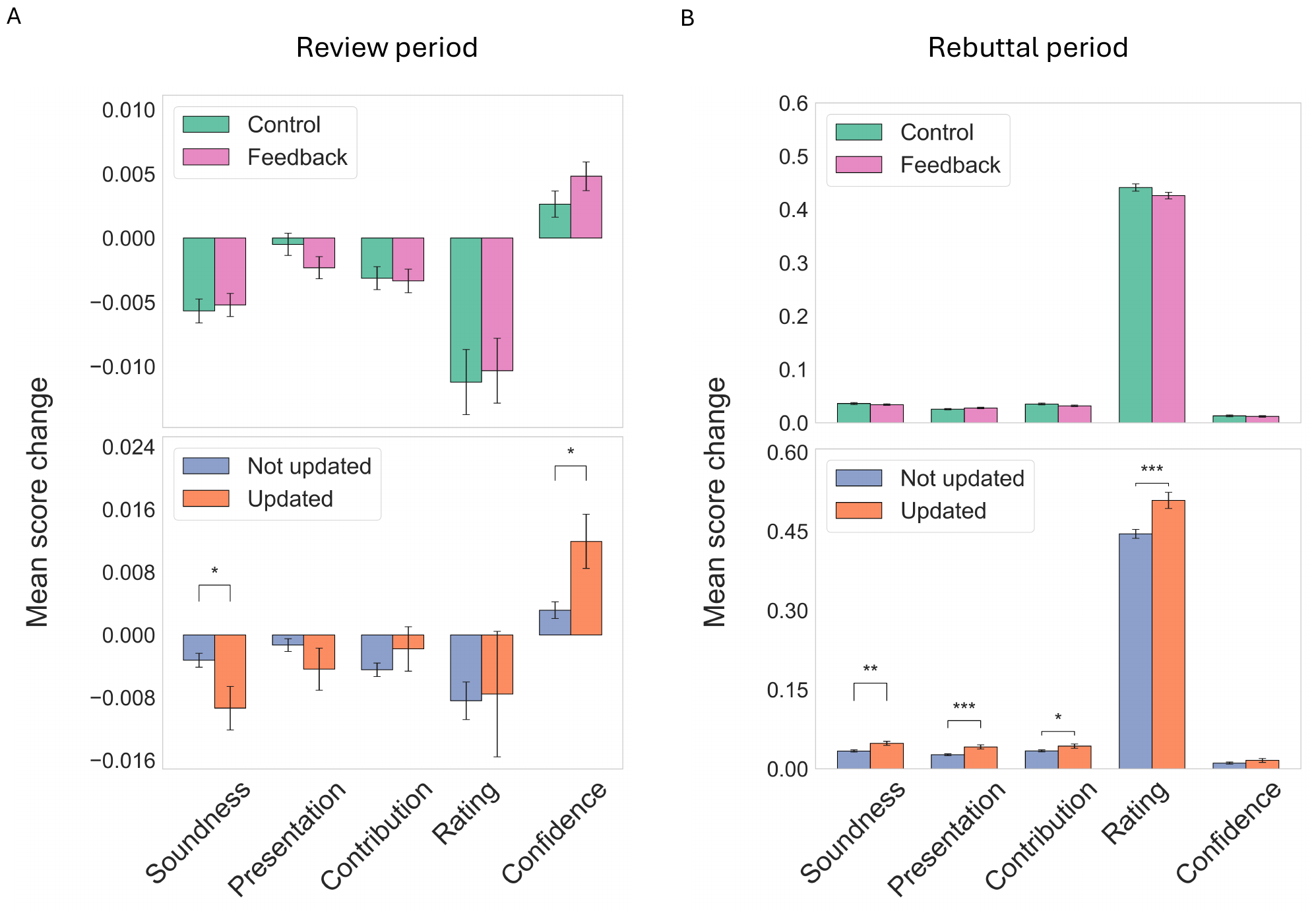}
    \caption{\textbf{(A) Review period score changes.} (Top) There is no significant difference in updating scores (measured between initial and pre-rebuttal reviews) between the feedback and control groups. (Bottom) Among reviewers who received feedback, those who updated their reviews were more likely to decrease soundness scores (p $\leq$ 0.05) and increase confidence scores (p $\leq$ 0.05) compared to those who did not update their reviews. \textbf{(B) Rebuttal period score changes.} (Top) There is no significant difference in updating scores (measured between pre- and post-rebuttal reviews) between the feedback and control groups. (Bottom) Among feedback recipients, reviewers who updated their reviews demonstrated significantly larger score increases across all metrics (soundness: $^{**}$p $\leq$ 0.01; presentation: $^{***}$p $\leq$ 0.001; contribution: $^{*}$p $\leq$ 0.05; rating: $^{***}$p $\leq$ 0.001) except confidence, compared to non-updaters.}
    \label{fig:supp_score_changes}
\end{figure}

In Figure \ref{fig:supp_score_changes}A, we examined the potential change in review scores (soundness, presentation, contribution, rating, and confidence) between the initial and modified reviews across the groups during the review period. We found that reviewers who were selected to receive feedback did not change their scores more than those in the control group (top panel). We also saw that of reviewers who received feedback, reviewers who updated their review were significantly more likely to decrease their soundness score and increase their confidence score at the end of the review period (before the rebuttal period began) compared to those who did not update their review. This suggests that reviewers who updated their reviews became more confident in their assessments.

In Figure \ref{fig:supp_score_changes}B, we conducted the same analysis during the rebuttal period. Similar to the review period, we found that reviewers who were selected to receive feedback did not change their scores more than those in the control group (top panel). Of reviewers who received feedback, those who updated their reviews significantly increased all scores except confidence compared to those who did not update their reviews. From this, we see that reviewers who updated their reviews were much more engaged in the rebuttal process.

\section{Incorporation model validation}
\label{sec:incorp_validation}

To test our incorporation model, we hand-labeled a test set of 222 feedback items (from 63 randomly chosen reviews that had been updated) as being incorporated into the updated review or not. We labeled 132 of those items as incorporated (59.5\%) and 90 as not (40.5\%). We then ran those 222 feedback items through the LLM pipeline and received a 92\% accuracy rate, with a false negative rate of 0.9\% and a false positive rate of 5.9\% (see Supplementary Figure \ref{fig:supp_conf_matrix}). Of the false positives, 8/13 were instances of human error where the labeler missed that the item was incorporated into the review, and the model accurately identified this incorporation. The remaining 5 false positives were due to subjectivity - the model reasoned that the reviewer partially incorporated the sentiments of the feedback, whereas the labeler did not view that as sufficient enough to count as incorporated. The two false negatives represent data points the labeler initially mislabeled and the model correctly labeled. This effectively gives us a false negative rate of 0\% and a false positive rate of 2.25\%, allowing us to be confident that our incorporation pipeline was highly accurate.

\begin{figure}[hbtp]
    \centering
    \includegraphics[width=\textwidth]{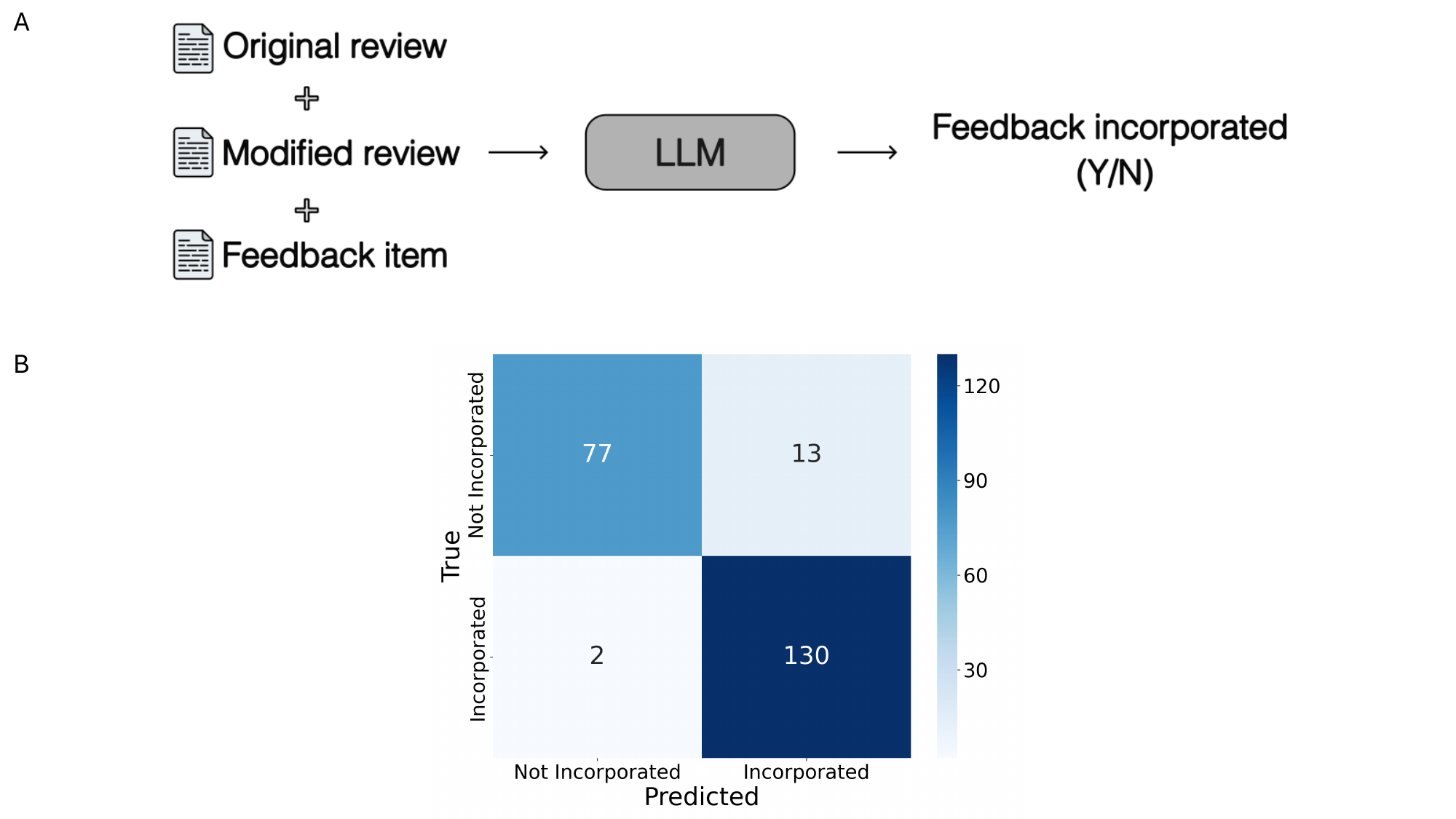}
    \caption{\textbf{(A) Incorporation model pipeline.} Given the original review text, modified review text, and individual feedback item, the LLM determined if the feedback was incorporated into the modified review or not. \textbf{(B) Model accuracy.} Our incorporation model successfully labeled 92\% of the test feedback items, where human annotators determined the ground truth labeling. Of the false positives, the majority were instances of human error where the model accurately identified the missed incorporation. All of the false negatives were instances of human error that the model caught.}
    \label{fig:supp_conf_matrix}
\end{figure}

\end{appendices}

\end{document}